\documentclass[number,sort&compress,5p]{elsarticle}

\usepackage[svgnames,x11names]{xcolor}

\usepackage[skip=8px,font=footnotesize]{caption} 
\usepackage[skip=8px,font=footnotesize]{subcaption}

\makeatletter
\renewcommand*\subcaption@label{%
  \caption@withoptargs}
\makeatother

\usepackage{paralist}
\usepackage{tikz}
\usetikzlibrary{calc}
\usetikzlibrary{matrix}
\usetikzlibrary{fit}

\def\fullthreeone{0.325}

\def\fullfourone{0.22}

\usepackage{mathtools} 
\usepackage{pgfplots}
\usepackage{pgfplotstable}
\usepgfplotslibrary{fillbetween} 

\usepackage{multirow}
\usepackage{lmodern}
\usepackage{wrapfig}
\usepackage{xspace} 

\usepackage{pifont} 

\usepackage{enumitem}
\usepackage{framed}
\usepackage{setspace}
\usepackage{textcomp}
\usepackage{multirow} 
\usepackage{placeins} 

\usepackage{tabularx}
\newcolumntype{Y}{>{\centering\arraybackslash}X}

\usepackage{array}
\usepackage{booktabs}
\usepackage{amssymb}
\usepackage[utf8]{inputenc}
\usepackage{graphicx}
\usepackage{enumerate}
\usepackage{color}
\usepackage{hyperref}

\hypersetup{
  colorlinks=true,
}
\usepackage[ocgcolorlinks]{ocgx2}
\usepackage[capitalise,nameinlink]{cleveref}
\crefname{figure}{Figure}{Figures}

\def\eg{{\em e.g.}}
\def\etal{{\em et al.}}
\def\ie{{\em i.e.}~}

\newcommand{\size}[1]{\left|#1\right|}
\def\SP{\mathcal{S}}
\def\GT{\mathcal{G}}
\newcommand{\RPG}[1]{$#1$-pixel grid}

\def\INFOMAP{\texttt{InfoMap}}
\def\LOUVAIN{\texttt{Louvain}}

\def\LP{\texttt{Label Propagation}}

\def\SEAL{Stutz~\etal~\cite{SHL18}}
\def\NGC{Nguyen~\etal~\cite{CD_NGC20}}

\def\ETPS{\textbf{ETPS}\xspace}

\def\SEEDS{\textbf{SEEDS}\xspace}
\def\SLIC{\textbf{SLIC}\xspace}
\def\TPS{\textbf{TPS}\xspace}

\def\EAMSR{\textbf{EAMS}~\cite{CM02}\xspace}
\def\ERSR{\textbf{ERS}~\cite{LTR+11}\xspace}
\def\ETPSR{\textbf{ETPS}~\cite{YBF+15}\xspace}
\def\FHR{\textbf{FH}~\cite{FH04}\xspace}
\def\QSR{\textbf{QS}~\cite{VS08}\xspace}
\def\SEEDSR{\textbf{SEEDS}~\cite{vdBBR+12}\xspace}
\def\SLICR{\textbf{SLIC}~\cite{ASS+12}\xspace}
\def\TPSR{\textbf{TPS}~\cite{FCD+14}\xspace}

\DeclareRobustCommand{\Kd}{%
    \ifmmode
        K_d
    \else
        $K_d$
    \fi
}
\def\UE{\text{UE}\xspace}

\def\Rec{\text{Rec}\xspace}

\def\CO{\text{CO}\xspace}
\def\EV{\text{EV}\xspace}

\newcommand{\FBSDS}{\href{https://www2.eecs.berkeley.edu/Research/Projects/CS/vision/grouping/resources.html}{Berkeley Segmentation Dataset 500}}
\newcommand{\FSBD}{\href{http://cs.nyu.edu/~silberman/datasets/nyu_depth_v2.html}{Stanford Background dataset}}
\newcommand{\FNYUV}{\href{http://cs.nyu.edu/~silberman/datasets/nyu_depth_v2.html}{NYU Depth Dataset V2}}
\newcommand{\FSUNRGBD}{\href{http://rgbd.cs.princeton.edu/}{Sun RGB-D dataset}}
\newcommand{\BSDS}{\texttt{BSDS500}}
\newcommand{\SBD}{\texttt{SBD}}
\newcommand{\NYUV}{\texttt{NYUV2}}
\newcommand{\SUNRGBD}{\texttt{SUNRGBD}}

\pgfplotsset{compat=1.18}
\pgfplotsset{
    LVN/.style={blue,solid,mark=diamond,mark options=solid},
    LP/.style={green,solid,mark=diamond,mark options=solid},
    INF/.style={red,solid,mark=diamond,mark options=solid},
    INF_TOP/.style={red,solid,mark=diamond,mark options=solid, mark size=0},
    INF_DOWN/.style={red,solid,mark=diamond,mark options=solid, mark size=0},
    EAMS/.style={gray,densely dashdotted,mark=square,mark options=solid},
    FH/.style={gray,dashed,mark=square,mark options=solid},
    QS/.style={gray,dotted,mark=heart,mark options={solid,fill=gray}},
    SLIC/.style={gray,dotted,mark=diamond,mark options=solid},
    ERS/.style={gray,dashed,mark=triangle*,mark options=solid},
    SEEDS/.style={black,dashed,mark=square*,mark options={solid,fill=black}},
    TPS/.style={gray,dashed,mark=o,mark options=solid},
    ETPS/.style={darkgray,densely dotted,mark=triangle*,mark options=solid},
    every tick label/.append style={
        font=\scriptsize,
    },
    every axis/.style={
        yticklabel style={
            /pgf/number format/fixed,
            /pgf/number format/precision=5
        },
        scaled y ticks=false,
        log ticks with fixed point,
		grid=both,
    },
    every axis plot/.append style={
		mark size=0.8,
    },
    every axis y label/.style={
		font=\scriptsize,
		at={(-0.175, 0.5)},
		rotate=90,
    },
    every axis x label/.style={
    	font=\scriptsize,
		at={(0.5, -0.175)},
    },
    EQBSDS500Rec/.style={
        height=5cm,
        width=6.5cm,
        ymin=0.5,
        ymax=1,
        xmin=180,
        xmax=6500,
        ylabel=\Rec,
        ytick={0.5,0.6,0.8,1.0},
        xtick={500, 1000, 3000, 6000},
    },
    EQBSDS500UE/.style={
        height=5cm,
        width=6.5cm,
        ymin=0.035,
        ymax=0.25,
        xmin=180,
        xmax=6500,
        ylabel=\UE,
        xtick={500, 1000, 3000, 6000},
        title=\BSDS,
    },
    EQBSDS500EV/.style={
        height=5cm,
        width=6.5cm,
        ymin=0.8,
        ymax=1,
        xmin=180,
        xmax=6500,
        ylabel=\EV,
        xtick={500, 1000, 3000, 6000},
    },
    EQNYUV2Rec/.style={
        height=5cm,
        width=6.5cm,
        ymin=0.5,
        ymax=1,
        xmin=180,
        xmax=6500,
        ylabel=\Rec,
        ytick={0.5,0.6,0.8,1.0},
        xtick={500, 1000, 3000, 6000},
    },
    EQNYUV2UE/.style={
        height=5cm,
        width=6.5cm,
        ymin=0.035,
        ymax=0.25,
        xmin=180,
        xmax=6500,
        ylabel=\UE,
        xtick={500, 1000, 3000, 6000},
        title=\NYUV,
    },
    EQNYUV2EV/.style={
        height=5cm,
        width=6.5cm,
        ymin=0.8,
        ymax=1,
        xmin=180,
        xmax=6500,
        ylabel=$\EV$,
        xtick={500, 1000, 3000, 6000},
    },
    EQSBDRec/.style={
        height=5cm,
        width=6.5cm,
        ymin=0.5,
        ymax=1,
        xmin=180,
        xmax=6500,
        ylabel=\Rec,
        ytick={0.5,0.6,0.8,1.0},
        xtick={500, 1000, 3000, 6000},
    },
    EQSBDUE/.style={
        height=5cm,
        width=6.5cm,
        ymin=0.035,
        ymax=0.25,
        xmin=180,
        xmax=6500,
        ylabel=\UE,
        xtick={500, 1000, 3000, 6000},
        title=\SBD,
    },
    EQSBDEV/.style={
        height=5cm,
        width=6.5cm,
        ymin=0.8,
        ymax=1,
        xmin=180,
        xmax=6500,
        ylabel=$\EV$,
        xtick={500, 1000, 3000, 6000},
    },
    EQSUNRGBDRec/.style={
        height=5cm,
        width=6.5cm,
        ymin=0.5,
        ymax=1,
        xmin=180,
        xmax=6500,
        ylabel=\Rec,
        ytick={0.5,0.6,0.8,1.0},
        xtick={500, 1000, 3000, 6000},
    },
    EQSUNRGBDUE/.style={
        height=5cm,
        width=6.5cm,
        ymin=0.035,
        ymax=0.25,
        xmin=180,
        xmax=6500,
        ylabel=\UE,
        xtick={500, 1000, 3000, 6000},
        title=\SUNRGBD,
    },
    EQSUNRGBDEV/.style={
        height=5cm,
        width=6.5cm,
        ymin=0.8,
        ymax=1,
        xmin=180,
        xmax=6500,
        ylabel=$\EV$,
        xtick={500, 1000, 3000, 6000},
    },
}

\begin{document}

\begin{frontmatter}

    \author{Anthony Perez}
    \journal{}
    \address{Univ. Orl\'eans, INSA Centre Val de Loire, LIFO EA 4022, F-45067 Orl\'eans, France}

\title{Superpixels algorithms through network community detection}

    \begin{abstract}
        Community detection is a powerful tool from complex networks analysis that finds applications in various research areas. 
        Roughtly speaking, it aims at grouping together nodes of a network that are densely connected while having few links with other groups. 
        Several image segmentation methods rely for instance on community detection algorithms as a black box in order to 
        compute \emph{undersegmentations}~\cite{CD_AMB14,CD_LBG+17,CD_HvGB+12,CD_BL16,CD_BAV11,CD_MEC19,CD_NGC20,CD_LW14}, \ie  
        a small number of regions that represent areas of interest of the image.  
        However, to the best of our knowledge, the efficiency of such an approach w.r.t. \emph{superpixels}, that 
        aim at representing the image at a smaller level while preserving as much as possible original information, has been neglected so far. 
        The only related work seems to be the one by Liu \etal~\cite{LDG+22} that developed a superpixels algorithm 
        using a so-called modularity maximization approach, leading to relevant results. We note that the algorithm used is a variant of a well-known  
        community detection algorithm that has however not been tested in a context other than image segmentation. 
        We follow this line of research by studying the efficiency of superpixels computed by state-of-the-art community detection algorithms 
        on a $4$-connected pixel graph, so-called \emph{pixel-grid}. We first detect communities on such a graph and then apply 
        a simple \emph{merging} procedure that allows to obtain the desired number of superpixels. 
        As we shall see, such methods result in the computation of relevant superpixels as emphasized by both qualitative and quantitative experiments, according 
        to different widely-used metrics based on ground-truth comparison or on superpixels only. 
        We observe that the choice of the community detection algorithm has a great impact on the number of communities and hence on the merging procedure. 
        Similarly, small variations on the pixel-grid may provide different results from both qualitative and quantitative viewpoints. 
        For the sake of completeness, we compare our results with those of several 
        state-of-the-art superpixels algorithms as computed by \SEAL. 
    \end{abstract}

\end{frontmatter}

\section{Introduction}
\label{sec:introduction}

In many real-life applications such as image segmentation or video analysis one may need to preprocess the 
image at hand in order to achieve great performances. One of the most natural such preprocessing is the computation of so-called 
\emph{superpixels} or \emph{oversegmentations} that represent the image at a smaller level while preserving as much as possible original information. 
Notable applications of superpixels include moving-object tracking, content-based image 
retrieval, biomedical imaging, indoor scene understanding, clothes parsing and convolutional neural networks. 
We refer the reader to recent comprehensive surveys for relevant references and more information on the topic~\cite{AE20,PVK23,SHL18}. 
We note that the literature mentions both the notion of superpixels and oversegmentation. 
According to \SEAL, the main difference lies in the possibility to 
control both the number of generated segments and their compactness, which leads to \emph{superpixels} when present 
and to \emph{oversegmentation} otherwise.  
As we shall see afterward community detection algorithms usually do not encompass an explicit way to control the number of segments but 
the merging procedure used does provide such a control. Moreover, the compactness can be adjusted by considering different graphs from the image. 
Hence the required characteristics for superpixels are fulfilled by this method: the segments form moreover a partition of the pixels, represent connected 
sets of pixels with great compactness and boundary adherence (see \SEAL). 
As we shall discuss later, computing superpixels is not efficient at the moment due to the implementation choices made for the sake of reproduciblity. 

\paragraph{Our contribution} 
We illustrate that the use of community detection algorithms on the most natural graph one can imagine, 
namely the pixel graph with $4$-connectivity (so-called \emph{pixel-grid}, 
see~\cref{fig:algorithm}) yields superpixels that achieve state-of-the-art results w.r.t. several metrics~\cite{SHL18}, both objective and subjective (\ie depending on 
ground-truth segmentations). We will also consider graphs with a given radius $r$, meaning that pixels are considered neighbors 
with all pixels at distance at most $r$. 
This work complements a previous analysis of Liu \etal~\cite{LDG+22} who used a variant of a well-known community detection algorithm on a 
similar graph to compute superpixels. We note here that such a variant has not been studied in a context other than image segmentation 
and that it seems to deeply rely on the particular structure of the pixel-grid. 
Our study compares three well-studied community detection algorithms on the same pixel-grid, namely \LP~\cite{EHR23}, \LOUVAIN~\cite{BGL+08} and \INFOMAP~\cite{RB08}.  
We emphasize that using different community detection algorithms on the same graph may lead to great differences 
on the outputted superpixels, mainly from the qualitative viewpoint.  
For the sake of reproducibility, we rely on the evaluation of state-of-the-art methods proposed by 
Stutz~\etal~\cite{SHL18}. As noticed by the authors, superpixel algorithms are often compared to other approaches with undisclosed 
or default parameter settings and with variating implementation of metrics. In particular, Stutz \etal~\cite[Appendix D]{SHL18} provide a thorough parameter 
optimization analysis. All the experiments proposed in our work rigourously reproduce the 
work of Stutz~\etal~\cite{SHL18} for which both implementation and plot files are available free of charge\footnote{\url{https://github.com/davidstutz/cviu2018-superpixels}}. 
The results presented illustrate the relevance of this approach compared to many state-of-the-art algorithms using different principles. 

\paragraph{Related work} There are a tremendous number of superpixels algorithm in the literature that rely on many different techniques. 
For instance, gradient-ascent and graph-based methods have 
been proposed for computing superpixels. In both cases some variations are possible but the main idea remains the same: starting with a primary grouping of pixels that is then refined until some convergence criterion is reached for the former; computing a graph based on pixels of the image for the latter (see~\cite{AE20}). 
We hereafter focus on graph-based methods due to their relevance with our work. 
In the last decades, many works took advantage of graph theoretic tools to compute segmentation algorithms. This is for instance the case of 
\emph{graph cuts} which rely on flow algorithms on the pixel-grid to compute hard segmentations of images~\cite{YM12}. 
Another notable use of the pixel-grid is the work of Felzenszwalb and Huttenlocher~\cite{FH04} who used minimum spanning trees algorithms to compute efficient segmentations. 
More recently, community detection algorithms have been used in several works as a tool to compute 
undersegmentations~\cite{CD_AMB14,CD_LBG+17,CD_HvGB+12,CD_BL16,CD_BAV11,CD_MEC19,CD_NGC20,CD_LW14}.   
In most cases the used algorithm relies either on a large 
graph or on a presegmented graph augmented with some features. 
For the former method, let us mention the work of \NGC{} who relied 
on the \LOUVAIN{} algorithm~\cite{BGL+08} on a pixel-grid of radius $20$ and on a merging procedure to obtain the sought segmentation. 
Regarding the latter approach, Mourchid \etal~\cite{CD_MEC19} first computed superpixels using Mean-Shift~\cite{CM02} as their initial segmentation and 
then used color and texture-based features to obtain their final segmentation through community detection algorithms.  
However, all aforementioned 
works aim at computing \emph{undersegmentations} and thus do not evaluate intermediate results that may yield oversegmentations meeting many 
requirements of superpixels. Note that this phenomenon is actually noticed in the work of \NGC{}.   
Very recently, Liu \etal~\cite{LDG+22} proposed a superpixels algorithm using a $8$-connected pixel-grid with radius $1$. 
The superpixels were detected using a greedy modularity maximization, a measure that quantifies the quality of a given community structure. 
Their algorithm heavily relies on the strong structural properties of pixel-grids and may thus not lead to relevant communities for other graphs. 
The final undersegmentation was also computed with a merging procedure. 

\paragraph{Outline} We begin by giving a general picture of the approach with a particular focus on community detection 
algorithms (\cref{sec:algorithm}).  
We next turn our attention to experimental results, both qualitative (\cref{subsec:qualitative}) and quantitative (\cref{subsec:quantitative}). 
Before doing so, we thoroughly describe datasets (\cref{subsec:datasets}), 
metrics (\cref{subsec:metrics}) and methods (\cref{subsec:methods}) used in our comparative studies. 
Finally, we give insights on the impact of the community detection algorithms used in \cref{sec:impact} and conclude this work by some 
perspectives \cref{sec:conclusion}. 

\section{Description of the framework}
\label{sec:algorithm}

\begin{figure*}[ht]
    \centering
	\begin{subfigure}[b]{\fullfourone\textwidth}
        \begin{center}
            Original image
        \end{center}
        \centering\includegraphics[height=2.25cm]{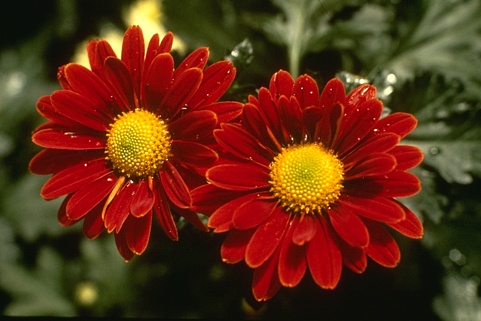}
    \end{subfigure}
	\begin{subfigure}[b]{\fullfourone\textwidth}
        \begin{center}
            Corresponding graph
        \end{center}
        \centering \includegraphics[height=2.25cm]{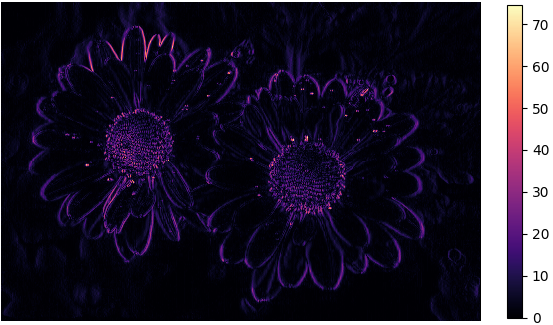}
    \end{subfigure}
	\begin{subfigure}[b]{\fullfourone\textwidth}
        \begin{center}
            Graph enhancement
        \end{center}
        \centering\includegraphics[height=2.25cm]{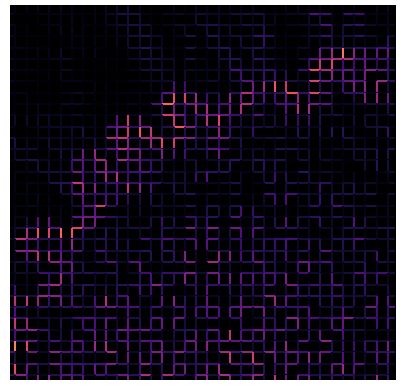}
    \end{subfigure}
	\begin{subfigure}[b]{\fullfourone\textwidth}
        \begin{center}
            Merged image
        \end{center}
        \centering \includegraphics[height=2.25cm]{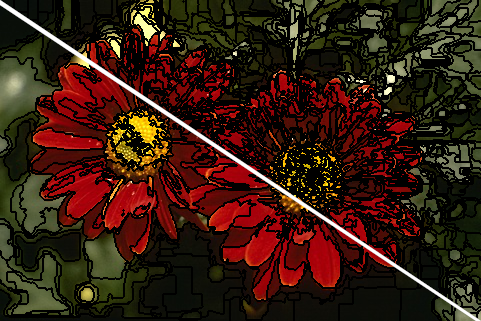}
    \end{subfigure}
    \caption{Different steps of the algorithm starting from an image from \BSDS. The lower part of the resulting image has $200$ superpixels while the upper part shows $1000$ regions. \label{fig:algorithm}}
\end{figure*}

Throughout the paper we consider $(n \times n)$-sized images with pixels $I = \{p_{1,1} \ldots, p_{n,n}\}$. 
As observed in previous works~\cite{CD_LW14,SHL18}, the \emph{L*a*b*} space is the closest to the human perception and is hence the one chosen for our study. 

\paragraph{Building the graph} A graph is a pair $G=(V,E)$ where $V$ denotes its vertex set and $E \subseteq [V]^2$ its edge set. 
Here $[V]^2$ denotes the set of all pairs of elements of $V$. 
Unless stated otherwise the considered 
graphs are simple (without self-loops nor multiedges), undirected and \emph{weighted}, meaning that every graph $G=(V,E)$ comes together with a weight 
function $\omega: E \rightarrow \mathbb{R}^+$. We consider as weight function a \emph{similarity measure} that will be described \cref{eq:weights}. 
We consider the simplest way to obtain a graph from an image, that is by considering 
pixels as neighbors using the $4$-connectivity model within a given distance. More precisely, given an integer $r \geqslant 1$, called \emph{radius},  
we define the \emph{\RPG{r}} graph of image $I$ as the graph 
$G_r = (V_r,E_r)$ where $V_r = I$ and there is an 
edge $pp'$ in $E_r$ if the corresponding pixels are both on the same row or the same column of $I$ and at distance at most $r$ from each other.  
(Obviously edges with out-of-bounds indices are not considered in this process.) 
Note that the $1$-pixel grid is simply the graph obtained from the direct $4$-connected neighborhood of every pixel. 
In this work we will consider values of $r$ ranging from $1$ to $10$, the best results being achieved for $r=5$ in most cases. 
For the sake of comparison, \NGC{} used a \RPG{20} to 
compute their undersegmentation, which implies that the used graph was significantly larger. 
Moreover, they removed edges whose similarity 
was below some fixed threshold $\rho$ but considered unweighted graphs. In our setting, we remove edges in a similar manner but we weight the 
remaining edges accordingly. We use a similarity measure based on the channel differences between pixels (or the mean of a region for a presegmented image), 
that is a Gaussian type radius basis function:
\begin{equation}
    \label{eq:weights}
    \omega(p,p') = \exp{\frac{-|p-p'|^2}{2\cdot \sigma^2}}
\end{equation}
where $\sigma$ is a parameter that defines how \emph{close} two regions must be for the corresponding edge weight to be significant.  
There are actually several graphs considered in the literature, some preserving all edges in a given radius (which corresponds to a threshold $\rho=0$) 
while other approaches preserve edges above a given threshold only. Moreover, some authors consider weighted graphs~\cite{LDG+22} while 
others use weight as a binary mask to remove or preserve unweighted edges~\cite{CD_NGC20}. As we shall see \cref{table:graphs} and \cref{sec:impact} these choices may 
have a great impact on pixel-grids and on computed communities. 

\paragraph{Computing communities} The next step for computing superpixels is to detect communities on the given \RPG{r}. 
Roughly speaking, a \emph{community structure} of a given graph is a partition\footnote{Let us mention that communities can 
    sometimes be defined as \emph{overlapping}, a feature that does not suit our purpose since we aim at computing 
partitions of pixels of the image.} of its vertex set such that every part is densely 
connected while there are few edges between two distinct parts. One may think of communities in social networks as a way to 
gather individuals that are similar according to some properties. Regarding image analysis, the intuition is that 
pixels that are similar (\eg{} regarding some color features) should be contained in the same community. 
There exist many community detection 
algorithms that sometimes exhibit different behaviors and that may hence encompass different properties of the graph~\cite{KH21}. Our study will use 
three algorithms for the sake of comparison, that we briefly describe hereafter. 

\begin{itemize}
    \item[-] \LP~\cite{EHR23} is an iterative algorithm that assigns labels to vertices, corresponding to communities. The initial step of the 
        algorithm labels every vertex $\{v_1, \ldots, v_n\}$ of an $n$-vertex graph with its corresponding index $1 \leqslant i \leqslant n$. Labels are then propagated 
        by considering vertices in a random order and giving to each vertex the majority label of its neighbors. The process stops when all vertices 
        are labeled with the majority label of their respective neighborhoods. A particular feature of this algorithm is that two consecutive runs 
        may end up with rather different community structures. 
    \item[-] \LOUVAIN~\cite{BGL+08} is an agglomerative algorithm that 
        greedily optimizes the so-called \emph{modularity} of the graph, a measure that qualifies a 
        given community structure. Roughly speaking, the 
        algorithm starts from an existing partition into 
        communities (\eg{} the singleton partition) and computes the \emph{gain} obtained by moving any vertex to a 
        different community. It stops when no valuable move 
        remains and then merges each community into a single vertex, 
        repeating this process until no valuable move is made. 
    \item[-] \INFOMAP{} relies on a flow-based and information theoretic method called the map equation~\cite{RB08}. 
        Quoting the work of Rosvall \etal~\cite{RAB09}: 
        \begin{quote}
            \emph{The map equation specifies the theoretical limit of how concisely 
            we can describe the trajectory of a given walker in the network. With such a random walker as a proxy 
            for real flow, minimizing the map equation over all possible network partitions reveals important aspects 
        of network structure with respect to the dynamics of the network. }
        \end{quote}
        Rosvall \etal~\cite{RAB09} moreover emphasize that methods based on the map equation and on modularity maximization 
        may yield really different community structures, making the study of those algorithms well-suited for our purpose. 
\end{itemize}

\begin{figure*}[ht]
    \centering
    \begin{subfigure}[b]{0.05\textwidth}
        ~
    \end{subfigure}
    \begin{subfigure}[b]{\fullfourone\textwidth}
		\centering{\small $r=1$}
	\end{subfigure}
    \begin{subfigure}[b]{\fullfourone\textwidth}
		\centering{\small $r=2$}
	\end{subfigure}
    \begin{subfigure}[b]{\fullfourone\textwidth}
		\centering{\small $r=5$}
    \end{subfigure}
    \begin{subfigure}[b]{\fullfourone\textwidth}
		\centering{\small $r=10$}
    \end{subfigure}\\[4px]
    \centering
    \begin{subfigure}[b]{0.05\textwidth}
		\rotatebox{90}{\small\hphantom{aaai}Image}
    \end{subfigure}
	\begin{subfigure}[b]{\fullfourone\textwidth}
        \centering\includegraphics[height=2.2cm]{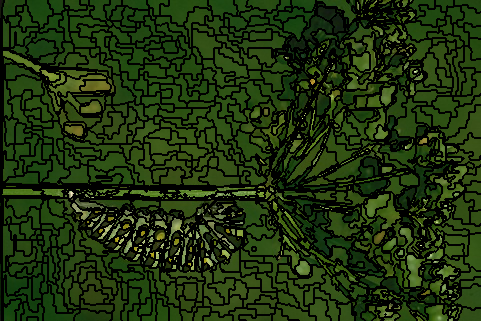}
    \end{subfigure}
	\begin{subfigure}[b]{\fullfourone\textwidth}
        \centering\includegraphics[height=2.2cm]{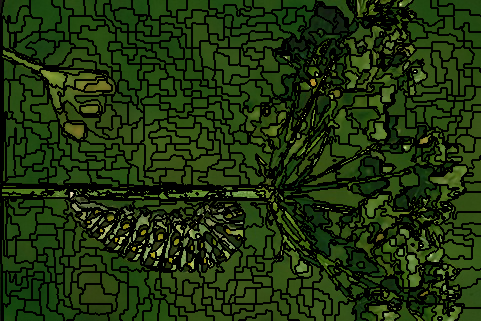}
    \end{subfigure}
	\begin{subfigure}[b]{\fullfourone\textwidth}
        \centering\includegraphics[height=2.2cm]{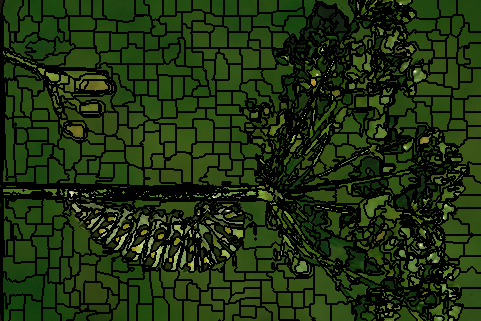}
    \end{subfigure}
	\begin{subfigure}[b]{\fullfourone\textwidth}
        \centering\includegraphics[height=2.2cm]{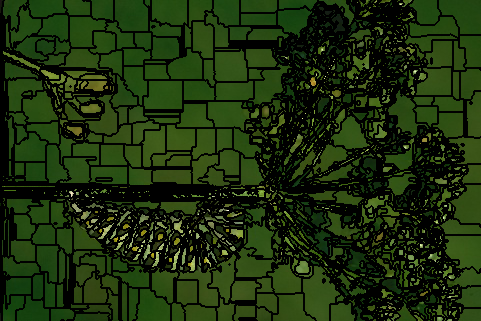}
    \end{subfigure}
    \caption{Impact of the radius for a sample image from \BSDS. Larger radius implies larger superpixels on background areas 
    with more small regions around objects and shapes. All images are computed with a threshold $\rho=0.98$ and $K=1000$. \label{fig:radius}}
\end{figure*}

\paragraph{Merging communities} 
The final step of the procedure is to reduce further the number of superpixels produced by the algorithms.  
A  similar procedure was used in several works~\cite{CD_NGC20,LDG+22} to produce image segmentation based on both the size and the similarity of superpixels. 
As noticed for instance by Liu \etal~\cite{LDG+22}, the sought number of superpixels $K$ 
can be explicitely managed by merging all communities with size smaller than $\frac{(n^2)}{K}$. 
Note that there are many criteria that can be considered for merging initial oversegmentations, such as the size of regions and 
the similarity between regions. 
We choose in this work to focus on a merging procedure based on the sizes of the communities, namely by merging any small enough region with its closest neighboring one w.r.t. 
similarity defined \cref{eq:weights} until the number of desired superpixels is reached. To that aim, we compute a \emph{Region Adjacency Graph} with radius $1$. 
We note that in order to avoid too many very small superpixels, 
w e begin by merging communities the size of which is less than $\frac{n^2}{10 \cdot K}$.  
Obviously, this approach allows to control the number of outputted regions, 
thus meeting the requirements of a superpixels algorithm. 
See \cref{fig:algorithm} for an example of the framework. 
Let us mention here that Liu \etal~\cite{LDG+22} chose a different approach in their work by 
merging small regions with the \emph{largest} neighboring one. 

\paragraph{Details of implementation} 
We conducted all experiments using \texttt{python3} with \href{https://scikit-image.org/}{scikit-image} for dealing with images,  
\href{https://networkx.org/}{networkx} and \href{https://networkit.github.io/}{networkit} for most graph-related operations and the module \href{https://mapequation.github.io/infomap/python/infomap.html}{Infomap}. 
Moreover, metrics are replicated from the \texttt{C++} \href{https://github.com/davidstutz/superpixel-benchmark}{implementation} of Stutz~\etal~\cite{SHL18}. 
Source code is available on a \href{https://github.com/anthonimes/superpixels-community-detection}{public github repository}. 
As mentioned in the introductory section, the aim of this study is to enlight the relevance of community detection to compute superpixels. 
We hence did not aim at achieving great performance in computing the segmentations. 
We will discuss this matter more thoroughly \cref{sec:conclusion}. 

\section{Experimental setup}
\label{sec:setup}

\subsection{Datasets} 
\label{subsec:datasets}

We now give a brief description of the datasets used in our experiments as well as references and links to retrieve the \emph{raw} datasets. 
Let us recall that we closely follow the work of Stutz \etal~\cite[Section 4]{SHL18} and use \href{https://github.com/davidstutz/superpixel-benchmark-data}{datasets} 
that have been preprocessed.  
Most of the following paragraphs are extracted from~\cite{SHL18}. 

\paragraph{\BSDS~\cite{AMF+11}} The \FBSDS{} is one of the earliest that has been used for superpixel algorithm evaluation. It contains 
$500$ images that come together with a set of at least $5$ ground truth for every image. We consider the $200$ images that 
are part of the \texttt{test} set. Following the work of \SEAL{} we choose for each image the ground truth providing the 
\emph{worst} score for a given metric and average such values over all images. 

\paragraph{\SBD~\cite{GFK09}} The \FSBD{} combines $715$ images from several datasets implying that the sizes, qualities 
and scenes are varying between images. 
The scenes tend to be more complex than those of the \BSDS{} dataset. The used dataset provided by 
\SEAL{} has been preprocessed to ensure connected segments in ground truth and contains $477$ randomly chosen images. 

\paragraph{\NYUV~\cite{SHK+12}} The \FNYUV{} contains $1449$ images. 
The labels provided by Silberman \etal~\cite{SHK+12} ensure connected segments and the data has been further 
preprocessed by \SEAL{} (following Ren and Bo~\cite{RB12}) to remove small unlabeled regions. 
Note that the provided ground truth is of lower quality than in the \BSDS{} dataset. 
The preprocessed dataset contains $399$ randomly chosen images.

\paragraph{\SUNRGBD~\cite{SLX15}} The \FSUNRGBD{} is made of  
$10335$ images combined from 
several datasets~\cite{JKJ+13,XOT13}, including \NYUV{} 
(which has here been excluded). Images have been acquired 
through several devices~\cite{SLX15} and ground truth has been preprocessed 
by \SEAL{} in a similar manner than for \NYUV.  
The preprocessed dataset contains $400$ images chosen at random.

\paragraph{Pixel-grids} \cref{table:graphs} provides some information about the pixel-grid obtained for different radii and threshold values on the \BSDS{} dataset. 
We use $G_r^\rho$ to denote a \RPG{r} where only edges with a weight greater than $\rho$ are preserved. 
Note that due to the threshold some vertices may become isolated and are thus not added to the graph. 
We also illustrate \cref{fig:radius} the impact the radius has on computed superpixels. We will discuss further these differences \cref{sec:results}. 

\begin{table}[hbt!]
    \centering
    \begin{tabular}{clllc}
        \toprule
        \multicolumn{1}{c}{~} & vertices & edges & weight & $\CO$ (\cref{table:metrics})\\
        \midrule
        \midrule
        $G_1^0$ & $154401$ & $308000$ & $254752$ & $0.31892$ \\ 
        $G_1^{0.98}$ & $97808$ & $117528$ & $116846$ & $0.19405$\\ 
        $G_2^0$ & $154401$ & $615198$ & $480182$ & $0.39695$ \\ 
        $G_2^{0.98}$ & $103765$ & $199446$ & $198228$ & $0.18930$ \\ 
        $G_5^{0}$ & $154401$ & $1531980$ & $1079130$ & $0.50955$ \\ 
        $G_5^{0.98}$ & $110143$ & $377524$ & $375030$ & $0.17446$ \\ 
        \bottomrule
    \end{tabular}
\caption{Number of vertices, edges and total edge weight for \RPG{r}s with $r$ in  
in $\{1,2,5\}$ and $\rho$ in $\{0, 0.98\}$. Values are rounded. Last column indicates 
the compactness for $1000$ superpixels on \BSDS. \label{table:graphs}} 
\end{table}

Finally, the last column gives the Compactness (see \cref{subsec:metrics}) on the \BSDS{} dataset for $K=1000$ superpixels. As one can observe, 
the best values are achieved for a radius $r=2$ \emph{and} a small threshold. This property actually holds regardless of the number of 
sought superpixels and thus allows to control the compactness of the superpixels. However, all other metrics presented \cref{table:metrics} 
have worse values for such combinations of radius and threshold. We hence choose not to include them in the remainder of our study. 
Nonetheless, let us mention that Stutz \etal~\cite[Fig. 8]{SHL18} provides Compactness values for all considered datasets. For $1000$ superpixels, 
the best value is achieved by \TPS{} with $0.56146$ while \SEEDS{} and \ETPS{} are the two lowest with $0.08952$ and $0.16260$, respectively. 
This enlightens that results provided by \INFOMAP{} are also relevant w.r.t. Compactness. 

\begin{table*}[t!]
    \begin{tabular}{c>{\small}l}
        \toprule
        Metric & \multicolumn{1}{c}{Notations} \\
        \midrule
        \midrule
        \multicolumn{2}{c}{\textbf{Ground truth-based}} \\
        ~ & ~ \\
        $\Rec(\GT,\SP) = \frac{TP(\GT,\SP)}{TP(\GT,\SP)+FN(\GT,\SP)}$ & $FN(\GT,\SP)$: number of false negative boundary pixels in $\SP$ w.r.t. $\GT$ \\ 
        ~                                                             & $TP(\GT,\SP)$: number of true positive boundary pixels in $\SP$ w.r.t. $\GT$ \\ 
        ~ & ~ \\
        $\UE(\GT,\SP) = \frac{1}{\size{I}} \cdot \sum_{G_i} \sum_{S_j \cap G_i \neq \emptyset} \min{\{\size{S_j \cap G_i},\size{S_j \setminus G_i}\}}$ & ~ \\
        ~ & ~ \\
        \midrule
        \midrule
        \multicolumn{2}{c}{\textbf{No ground truth}} \\
        ~ & ~ \\
        $\EV(\SP) = \frac{\sum_{S_j} \size{S_j} \cdot (\mu(S_j) - \mu(I))^2}{\sum_{1 \leqslant i \leqslant j \leqslant n} (I(p_{i,j}) - \mu(I))^2}$ & $\mu(\cdot)$: mean color of its argument \\
        ~ & ~ \\
        $\CO(\SP) = \frac{1}{\size{I}} \cdot \sum_{S_j} \size{S_j} \cdot \frac{4\Pi \cdot A(S_j)}{P(S_j)}$ & $A(S_j)$: area of superpixel $S_j$ \\ 
        ~                                                                                                  & $P(S_j)$: perimeter of superpixels $S_j$ \\       
        \bottomrule 
    \end{tabular}
    \caption{Presentation of the metrics considered in this work together with needed notations. \label{table:metrics}}
\end{table*}

\subsection{Evaluation metrics} 
\label{subsec:metrics}

There are many metrics that can be used to assess the quality of superpixels algorithms, mostly relying on a ground truth 
segmentation of the image at hand. In this work we choose four metrics that have been introduced in various articles~\cite{NP12,LTR+11,SFS12,MPW+08}. 
Our choice is here again guided by Stutz \etal~\cite[Section 5]{SHL18} who give a detailed analysis of many metrics and ultimately focus on these ones. 
In the remaining of this section and in a slight abuse of notation we let $I(p_{i,j})$,  
$1 \leqslant i \leqslant j \leqslant n$ denote the intensity of pixel $p_{i,j}$.   
We moreover let $\SP = \{S_1, \ldots, S_p\}$ and $\GT = \{G_1, \ldots, G_t\}$ 
be partitions of a same image $I$ into superpixels and ground truth segmentation, respectively. All 
metrics are summarized \cref{table:metrics}. We give hereafter a brief overview of such metrics as presented by \SEAL.  \\

\emph{Boundary recall} (\Rec) is widely used to assess the quality of a superpixel segmentation, and more precisely boundary adherence given ground truth. 
\emph{Under-segmentation error} ($\UE$) aims at measuring the overlap of superpixels with multiple, nearby ground truth segments.  
We note that the original formulation for Under-Segmentation Error was given by Levinshtein \etal~\cite{LSK09} as follows:
\begin{equation*}
    \UE_{Levin}(\GT,\SP) = \frac{1}{\size{\GT}} \cdot \sum_{G_i} \frac{\big (\sum_{S_j \cap G_i \neq \emptyset} \size{S_j} \big ) - \size{G_i}}{\size{G_i}}
\end{equation*}

However, as observed for instance in~\cite{ASS+12,NP12} such a definition penalizes superpixels overlapping only 
slightly with neighboring ground truth segments and is not constrained to lie in $[0,1]$. 
The adapted version chosen by \SEAL{} has been proposed by Neubert and Protzel~\cite{NP12}.  
\emph{Explained variation} ($\EV$)~\cite{MPW+08} quantifies the quality of a superpixel segmentation 
by assessing boundary adherence without relying on human annotations. 
High \EV{} scores means better superpixels explanation of the variation of the image. 
\emph{Compactness} (\CO) has been introduced by Schick \etal~\cite{SFS12} to evaluate the compactness of superpixels. 
Compactness compares the area of a superpixel $S_j$ with the area of 
a circle with the same perimeter $P(S_j)$. High \CO{} value means better compactness. 

\subsection{Methods for comparison} 
\label{subsec:methods}

We now briefly describe the state-of-the art 
methods used in our experiments. We based our choices on the work of Stutz \etal~\cite{SHL18} which highlighted the efficiency of chosen methods w.r.t. metrics defined \cref{subsec:metrics}. In order to present a comparative study as meaningful as possible, we selected a couple of the best methods for different superpixels paradigms, namely 
density, graph, path, clustering and energy optimization based methods. The presentation of such methods is inspired by~\cite[Section 3]{SHL18}. 

\paragraph{Density-based} Such algorithms \emph{perform mode-seeking in a computed density image and usually do not offer control on the number 
of superpixels or their compactness}~\cite{SHL18}. Yet tuning the parameters can allow to achieve such objectives. We compare the presented method to \EAMSR{} and \QSR. 

\paragraph{Graph-based} The common feature of these methods is to extract a graph from the image at hand and then to use graph-theoretic tools to compute 
superpixels. Many tools have been used, such as network flows (graph cuts~\cite{YM12}) and minimum spanning trees~\cite{FH04}. Hence the main difference 
lies in the method used to compute the superpixels, the graph usually being based on a pixel (dis)similarity measure. We compare the presented method to \FHR{} and \ERSR. 

\paragraph{Path-based} Methods based on path detection compute superpixels by creating paths between seeds respecting some criteria. We compare the presented method to the 
\TPSR{} algorithm, which relies on edge detection. 

\paragraph{Clustering-based} These algorithms are inspired by clustering methods such as $k$-means and are close to the approach we consider 
here. Like for path-based methods, clustering-based superpixels algorithms use seeds and several image-related information as 
features to group pixels together. As noted in \SEAL, the resulting superpixels may be disconnected and further postprocessing is needed 
to ensure connectivity. We compare the presented method to \SLICR. 

\paragraph{Energy optimization-based} Starting from a regular grid, pixels are then exchanged with neighborhing superpixels w.r.t. 
some energy function. These methods achieve high accuracy and efficiency. We compare the presented method to \SEEDSR{} and \ETPSR. 

\section{Experimental results}
\label{sec:results}

We focus in this section on superpixels obtained using \INFOMAP, which provides the best 
trade-off w.r.t. the quality of both qualitative and quantitative comparisons. 
We will see \cref{sec:impact} that using different community detection algorithms such as \LP{} or \LOUVAIN{} 
may lead to a bit worse quantitative results with a significant qualitative difference. 
Moreover, one advantage of such an algorithm compared to \LP{} or \LOUVAIN{} is that it gives consistent community structures 
from one image to another, allowing a better control on the number of superpixels in the merging procedure.  

\paragraph{Parameters} Besides the parameters specific to the chosen community detection algorithms this approach does not require 
a lot of parameter tuning. More precisely, one needs to fix values for the radius of the pixel-grid $r$, the threshold $\rho$ relative to 
\cref{eq:weights} to decide which edges to preserve, the number of desired superpixels and the parameter $\alpha$ in \cref{eq:weights}. 
The latter has been experimentally chosen as $\alpha=125$. 
We recall \cref{table:graphs} for the size of the pixel-grid according to both radius and threshold values. 
Our experiments show that choosing a combination of $r=5$ and $\rho=0.98$ 
provides relevant superpixels while maintaining a graph of reasonable size. We will present \cref{sec:impact} some results using different radii  
and observe that their performances are a bit worse. Similarly, metrics computed with smaller threshold values were almost always 
significantly worse and are thus not reported here, with the notable exception of Compactness which can be improved using smaller values for $\rho$ 
(the better being $0$, meaning not filtering the pixel-grid, recall \cref{table:graphs}). 
Regarding the merging procedure the only parameter needed is the desired number of superpixels, which will vary from $200$ to 
$5000$ with different increasing steps: $200$ up to $1000$ and $500$ up to $2500$. We also compute $5000$ superpixels. 

\subsection{Quantitative comparison}
\label{subsec:quantitative}

We first turn our attention to quantitative comparison with methods mentioned~\cref{subsec:methods} on datasets described~\cref{subsec:datasets}. 
We compare our results with results extracted from the work of Stutz~\etal~\cite{SHL18}.  
Recall that we compute the metrics on superpixels generated using \INFOMAP{} on a \RPG{5} and with $\rho=0.98$ as threshold for all datasets, which provide the most relevant results. 
However, we will see \cref{fig:quantitative_metrics} that small variations of the radius may have an impact on the computed metrics. 

\begin{figure*}
    \centering
    \input{plots/bsds500} 
    \input{plots/nyuv2}
    \input{plots/sbd}
    \input{plots/sunrgbd}
    \centering\includegraphics[scale=1]{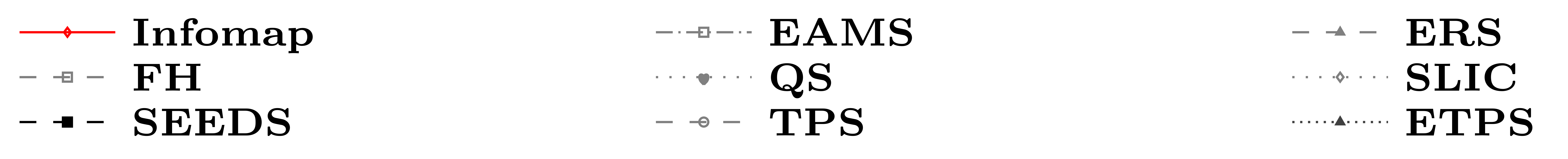}
    \caption{Quantitative comparison on datasets described~\cref{subsec:datasets}. Values for other methods are reported from \SEAL. 
    The results presented for this framework use a threshold value $\rho=0.98$ and a radius $r=5$. \label{fig:quantitative}} 
\end{figure*}

As one can see \cref{fig:quantitative}, this approach produces relevant results for all metrics. 
The recall values lie in the top three methods in any dataset. 
A similar observation holds for \UE{} values while the best \EV{} is achieved for a large number of superpixels.  
We also display standard deviation on metrics for all datasets. For the sake of readability, we do not 
display standard deviations for state-of-the-art methods. Such values can be found in~\cite[Fig. 9]{SHL18} 
and indicate that \SEEDS{} and \ETPS{} tend to have smaller standard deviations than \INFOMAP{} for $\Rec$. 
On the other hand, the displayed values for $\UE$ and $\EV$ are rather similar than the ones in~\cite[Fig. 9]{SHL18}. 
We note that for the \SBD{} dataset the mean number of superpixels when $K=5000$ is actually $4984$. This comes from the 
fact that for a few images, \INFOMAP{} computed a small number of communities and thus the method 
does not produce the exact number of superpixels. This is the case for exactly five images, and the number of 
superpixels produced for three of them is less than $4000$. 

\subsection{Qualitative comparison}
\label{subsec:qualitative}

\begin{figure*}[hbt!]
    \centering
    \begin{subfigure}[b]{0.05\textwidth}
		\rotatebox{90}{\small\hphantom{aaai}Image}
    \end{subfigure}
	\begin{subfigure}[b]{\fullfourone\textwidth}
        \begin{center}
            \BSDS
        \end{center}
        \centering\includegraphics[height=2.2cm]{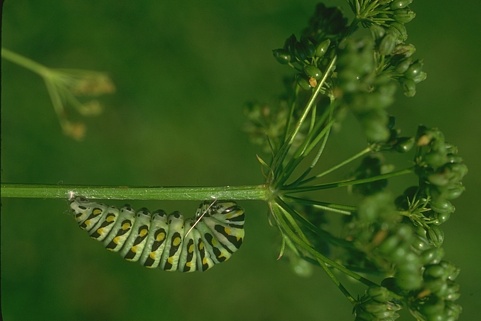}
    \end{subfigure}
	\begin{subfigure}[b]{\fullfourone\textwidth}
        \begin{center}
            \SBD
        \end{center}
        \centering\includegraphics[height=2.2cm]{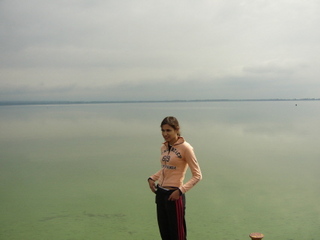}
    \end{subfigure}
	\begin{subfigure}[b]{\fullfourone\textwidth}
        \begin{center}
            \NYUV
        \end{center}
        \centering\includegraphics[height=2.2cm]{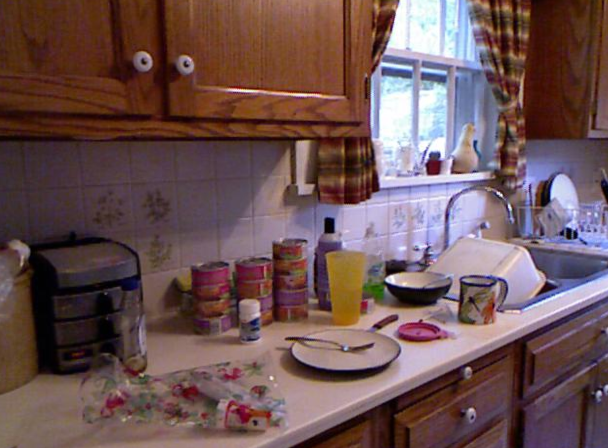}
    \end{subfigure}
	\begin{subfigure}[b]{\fullfourone\textwidth}
        \begin{center}
            \SUNRGBD
        \end{center}
        \centering\includegraphics[height=2.2cm]{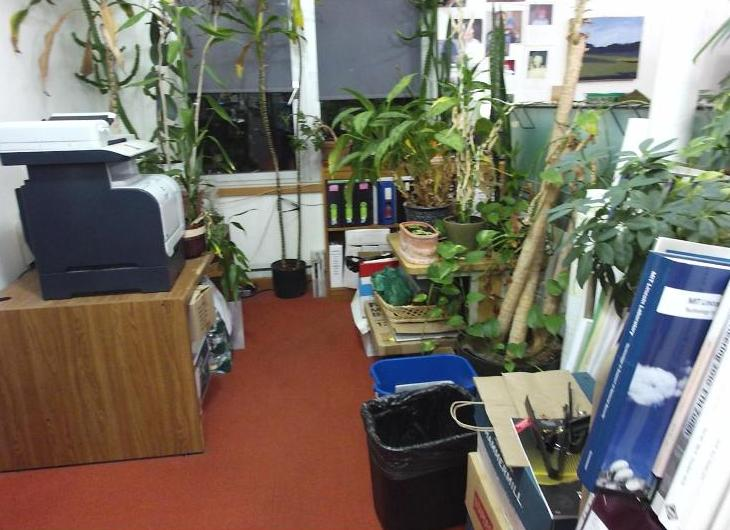}
    \end{subfigure}\\[4px]
    \centering
    \begin{subfigure}[b]{0.05\textwidth}
		\rotatebox{90}{\small\hphantom{aaai}\SEEDS}
	\end{subfigure}
	\begin{subfigure}[b]{\fullfourone\textwidth}
        \centering\includegraphics[height=2.25cm]{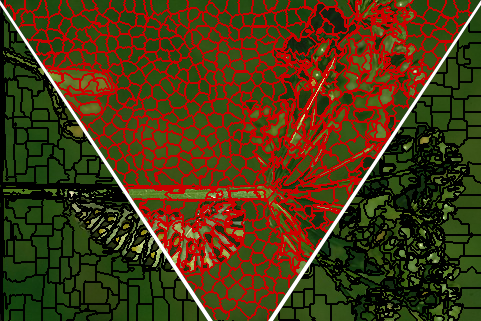}
    \end{subfigure}
	\begin{subfigure}[b]{\fullfourone\textwidth}
        \centering\includegraphics[height=2.25cm]{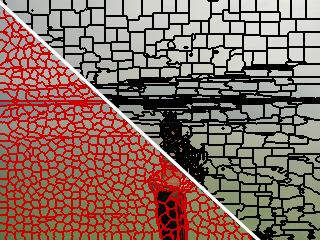}
    \end{subfigure}
	\begin{subfigure}[b]{\fullfourone\textwidth}
        \centering\includegraphics[height=2.25cm]{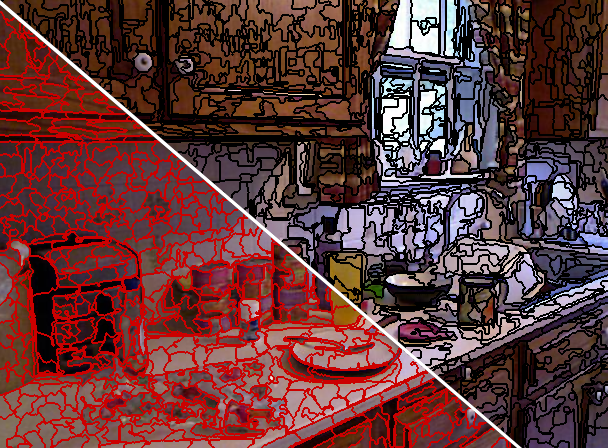}
    \end{subfigure}
	\begin{subfigure}[b]{\fullfourone\textwidth}
        \centering\includegraphics[height=2.25cm]{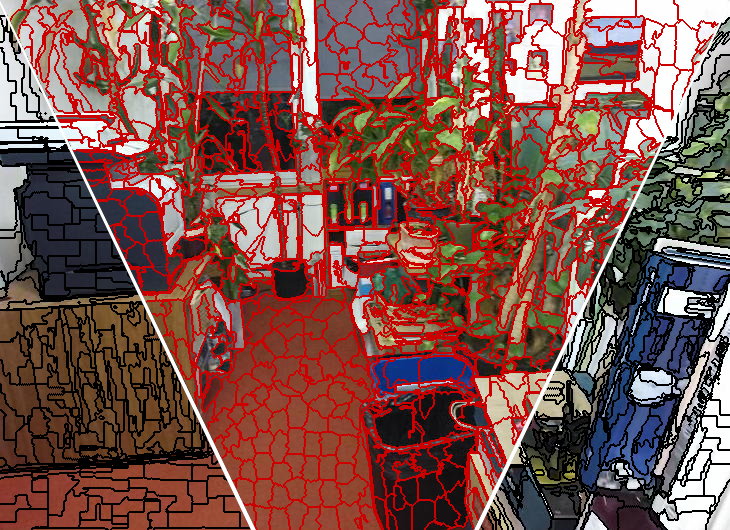}
    \end{subfigure}\\[4px]
    \centering
    \begin{subfigure}[b]{0.05\textwidth}
		\rotatebox{90}{\small\hphantom{aaai}\ETPS}
	\end{subfigure}
	\begin{subfigure}[b]{\fullfourone\textwidth}
        \centering\includegraphics[height=2.25cm]{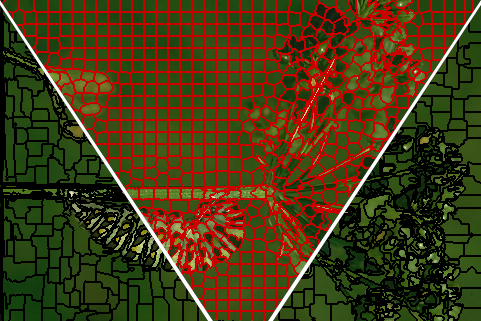}
    \end{subfigure}
	\begin{subfigure}[b]{\fullfourone\textwidth}
        \centering\includegraphics[height=2.25cm]{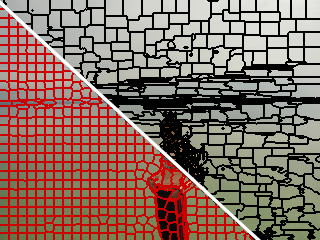}
    \end{subfigure}
	\begin{subfigure}[b]{\fullfourone\textwidth}
        \centering\includegraphics[height=2.25cm]{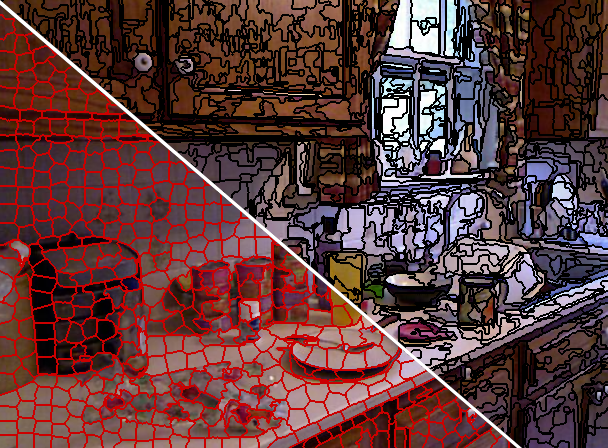}
    \end{subfigure}
	\begin{subfigure}[b]{\fullfourone\textwidth}
        \centering\includegraphics[height=2.25cm]{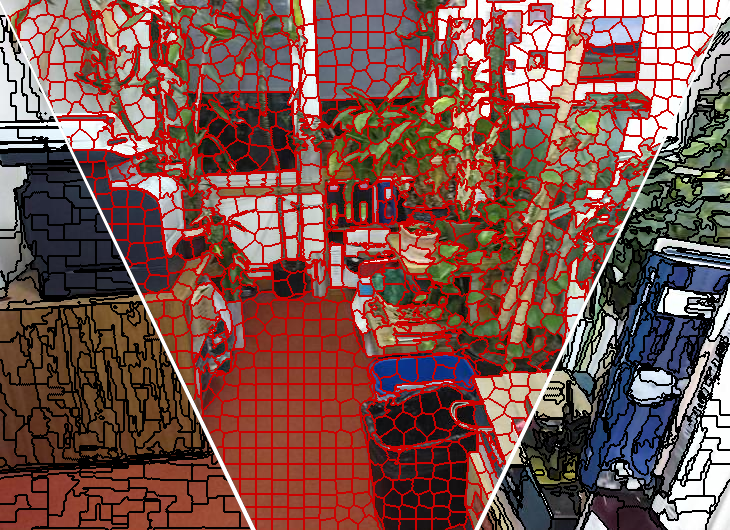}
    \end{subfigure}\\[4px]
    \centering
    \begin{subfigure}[b]{0.05\textwidth}
		\rotatebox{90}{\small\hphantom{aaai}\SLIC}
	\end{subfigure}
	\begin{subfigure}[b]{\fullfourone\textwidth}
        \centering\includegraphics[height=2.25cm]{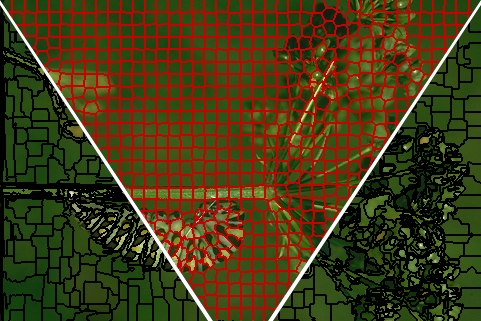}
    \end{subfigure}
	\begin{subfigure}[b]{\fullfourone\textwidth}
        \centering\includegraphics[height=2.25cm]{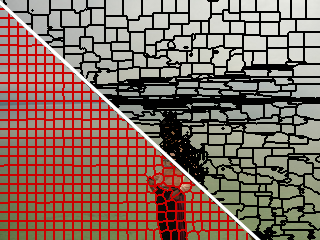}
    \end{subfigure}
	\begin{subfigure}[b]{\fullfourone\textwidth}
        \centering\includegraphics[height=2.25cm]{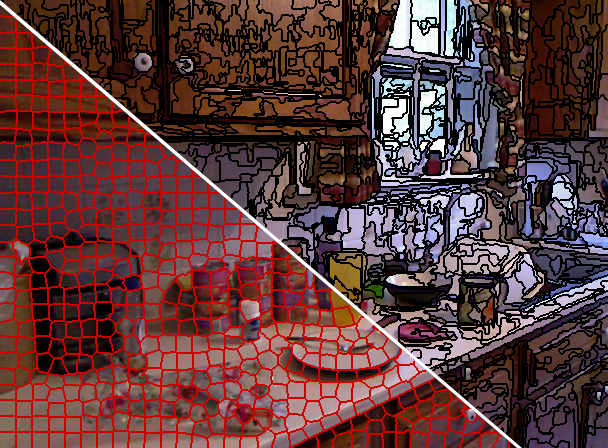}
    \end{subfigure}
	\begin{subfigure}[b]{\fullfourone\textwidth}
        \centering\includegraphics[height=2.25cm]{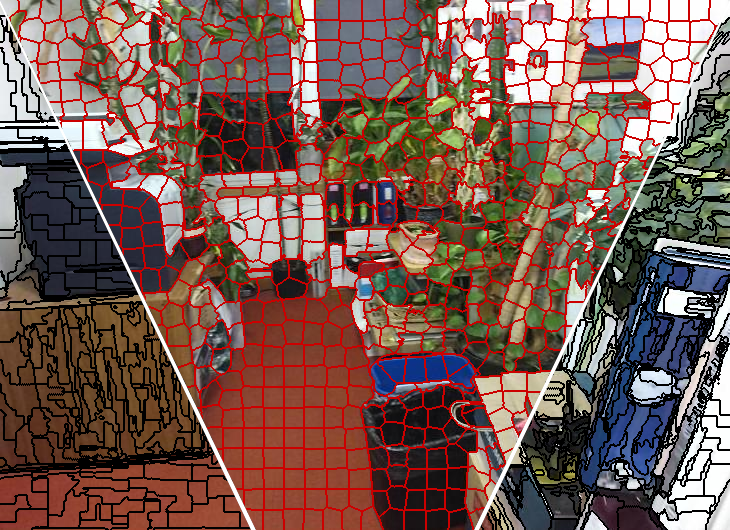}
    \end{subfigure}
    \caption{Qualitative comparison with selected methods, namely \SEEDS, \SLIC{} and \ETPS{} on selected datasets with a number of superpixels of $K=1000$. \label{fig:qualitative}}
\end{figure*}

We now give some examples from datasets that illustrate the superpixels achieved by this method compared to some approaches
described~\cref{subsec:methods}. \cref{fig:qualitative} emphasizes that the computed superpixels exhibit regularity 
and compactness on background areas and are less regular around objects. They show some similarities with superpixels computed by \SEEDS{} and \ETPS,  
methods that present the best result w.r.t. chosen metrics as seen \cref{subsec:quantitative}. On the other hand, their regularity and compactness 
is not as clearly defined as \SLIC. 
Note however that as shown \cref{table:graphs} and \cref{fig:radius} the above properties can be adjusted by using different radius and threshold values. 
Hence, depending on the image at hand a simple adjustment of these values may lead to significant differences between superpixels.  

\section{The impact of community detection}  
\label{sec:impact}

In this section we discuss the impact the choice of the community detection algorithm may have on the computed superpixels. 
We limit our study to the \BSDS{} dataset which provides meaningful insights. 
We begin by showing \cref{table:communities_stats} the number of communities computed by each algorithm for 
radii $r \in \{2,5\}$ and threshold $\rho \in \{0,0.98\}$ used for building the pixel-grid. 
The difference in the number of computed communities along with their sizes emphasizes a difference in the regularity and the sizes of 
said communities. This has a great impact on the merging procedure that provides the final superpixels. 

\begin{table*}
    \begin{tabular}{l>{$}c<{$}>{\footnotesize\color{gray}($}c<{$)}||>{$}c<{$}>{\footnotesize\color{gray}($}c<{$)}||>{$}c<{$}>{\footnotesize\color{gray}($}c<{$)}}
        \toprule
        $G_2^0$ & \multicolumn{2}{c}{Communities} & \multicolumn{2}{c}{Max. size} & \multicolumn{2}{c}{Min. size} \\
        & mean & std & mean & std & mean & std \\
        \midrule
        \texttt{LP} & 5760 & 3228 & 2579 & 1231 & 1.2 & 0.39 \\
        \LOUVAIN & 79 & 29 & 5840 & 1653 & 97 & 234 \\
        \INFOMAP & 2780 & 634 & 206 & 25 & 2 & 0.34 \\
        \bottomrule
    \end{tabular}
    \begin{tabular}{l>{$}c<{$}>{\footnotesize\color{gray}($}c<{$)}||>{$}c<{$}>{\footnotesize\color{gray}($}c<{$)}||>{$}c<{$}>{\footnotesize\color{gray}($}c<{$)}}
        \toprule
        $G_2^{0.98}$ & \multicolumn{2}{c}{\#Communities} & \multicolumn{2}{c}{Max. size} & \multicolumn{2}{c}{Min. size} \\
        & mean & std & mean & std & mean & std \\
        \midrule
        \texttt{LP} & 72359 & 23599 & 726 & 635 & 1 & 0 \\
        \LOUVAIN & 63230 & 25616 & 2126 & 1266 & 1 & 0\\
        \INFOMAP & 65374 & 24873 & 133 & 36 & 1 & 0\\
        \bottomrule
    \end{tabular}\\[4px]
    \begin{tabular}{l>{$}c<{$}>{\footnotesize\color{gray}($}c<{$)}||>{$}c<{$}>{\footnotesize\color{gray}($}c<{$)}||>{$}c<{$}>{\footnotesize\color{gray}($}c<{$)}}
        \toprule
        $G_5^{0}$ & \multicolumn{2}{c}{\#Communities} & \multicolumn{2}{c}{Max. size} & \multicolumn{2}{c}{Min. size} \\
        & mean & std & mean & std & mean & std \\
        \midrule
        \texttt{LP} & 1924 & 1308 & 10940 & 6120 & 1.79 & 0.41 \\
        \LOUVAIN & 46 & 19 & 10759 & 3786 & 180 & 563 \\
        \INFOMAP & 1054 & 335 & 684 & 93 & 2.06 & 0.31 \\
        \bottomrule
    \end{tabular}
    \begin{tabular}{l>{$}c<{$}>{\footnotesize\color{gray}($}c<{$)}||>{$}c<{$}>{\footnotesize\color{gray}($}c<{$)}||>{$}c<{$}>{\footnotesize\color{gray}($}c<{$)}}
        \toprule
        $G_5^{0.98}$ & \multicolumn{2}{c}{\#Communities} & \multicolumn{2}{c}{Max. size} & \multicolumn{2}{c}{Min. size} \\
        & mean & std & mean & std & mean & std \\
        \midrule
        \texttt{LP} & 63909 & 23472 & 2053 & 1890 & 1 & 0 \\
        \LOUVAIN & 54998 & 24671 & 3490 & 2194 & 1 & 0\\
        \INFOMAP & 57176 & 24175 & 371 & 115 & 1 & 0\\
        \bottomrule
    \end{tabular}
    \caption{Statistics of the communities computed by algorithms on the \BSDS{} dataset. The number of communities as well as the maximum and minimum sizes are reported. 
    For all values we show mean and standard deviation. \label{table:communities_stats}}
\end{table*}

\paragraph{Quantitative difference between algorithms}

\pgfplotsset{
    LVN/.style={blue,solid,mark=diamond,mark options=solid},
    LP/.style={green,solid,mark=diamond,mark options=solid},
    INF/.style={red,solid,mark=diamond,mark options=solid},
    every tick label/.append style={
        font=\scriptsize,
    },
    every axis/.style={
        yticklabel style={
            /pgf/number format/fixed,
            /pgf/number format/precision=5
        },
        scaled y ticks=false,
        log ticks with fixed point,
		grid=both,
    },
    every axis plot/.append style={
		mark size=1.2,
    },
    every axis y label/.style={
		font=\scriptsize,
		at={(-0.175, 0.5)},
		rotate=90,
    },
    every axis x label/.style={
    	font=\scriptsize,
		at={(0.5, -0.175)},
    },
    EQBSDS500Rec/.style={
        height=4cm,
        width=6.5cm,
        ymin=0.85,
        ymax=1,
        xmin=1,
        xmax=10,
        ylabel=\Rec,
        xtick={1, 2, 5, 10},
        title=\BSDS,
    },
    EQBSDS500UE/.style={
        height=4cm,
        width=6.5cm,
        ymin=0.035,
        ymax=0.15,
        xmin=1,
        xmax=10,
        ylabel=\UE,
        xtick={1, 2, 5, 10},
    },
    EQBSDS500EV/.style={
        height=4cm,
        width=6.5cm,
        ymin=0.9,
        ymax=0.975,
        xmin=1,
        xmax=10,
        ylabel=\EV,
        xtick={1, 2, 5, 10},
    },
}

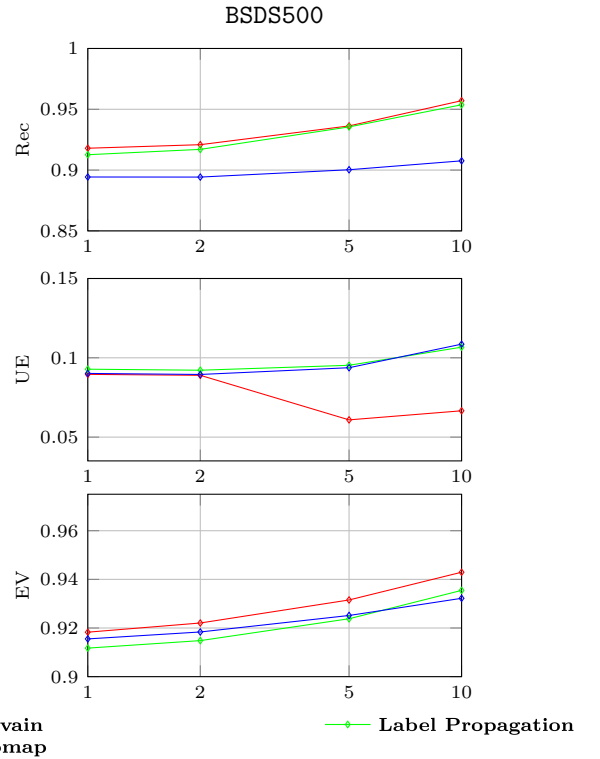
\begin{figure}
    \centering
    \begin{minipage}[b]{\fullthreeone\textwidth}
	\begin{tikzpicture}
		\begin{axis}[EQBSDS500Rec,xmode=log]

			\addplot[INF] coordinates{
				(1,0.9179678426729173)
				(2,0.9208822009937906)
				(5,0.9363091060197618)
				(10,0.957061981372708)
			};

			\addplot[LP] coordinates{
				(1,0.9126320947449497)
				(2,0.9170283440505537)
				(5,0.935501896038542)
				(10,0.9536876381530519)
			};

			\addplot[LVN] coordinates{
				(1,0.8943076075132724)
				(2,0.8942474624210162)
				(5,0.9002797671979194)
				(10,0.9075964147810606)
			};

		\end{axis}
	\end{tikzpicture}
\end{minipage}
\begin{minipage}[b]{\fullthreeone\textwidth}
	\begin{tikzpicture}
		\begin{axis}[EQBSDS500UE,xmode=log]

			\addplot[INF] coordinates{
				(1,0.08955783965129761)
				(2,0.08896347821581467)
				(5,0.0608714477546253)
				(10,0.06659159702734468)
			};

			\addplot[LP] coordinates{
				(1,0.0928163030032189)
				(2,0.09217297815428656)
				(5,0.09528357976956109)
				(10,0.10669671828550337)
			};

			\addplot[LVN] coordinates{
				(1,0.09015106767443214)
				(2,0.08949631155238627)
				(5,0.09375172440593005)
				(10,0.10859906995421012)
			};

		\end{axis}
	\end{tikzpicture}
\end{minipage}
\begin{minipage}[b]{\fullthreeone\textwidth}
	\begin{tikzpicture}
		\begin{axis}[EQBSDS500EV,xmode=log]

			\addplot[INF] coordinates{
				(1,0.9182811363443878)
				(2,0.9220584397978144)
				(5,0.9315275567274367)
				(10,0.9429352377749939)
			};

			\addplot[LP] coordinates{
				(1,0.9117303714887608)
				(2,0.914816085998395)
				(5,0.9238052782235172)
				(10,0.9354669469527446)
			};

			\addplot[LVN] coordinates{
				(1,0.9155141591435744)
				(2,0.9183684884692772)
				(5,0.9251492395832243)
				(10,0.9322131538875759)
			};

		\end{axis}
	\end{tikzpicture}
\end{minipage}
    \centering\includegraphics[scale=1]{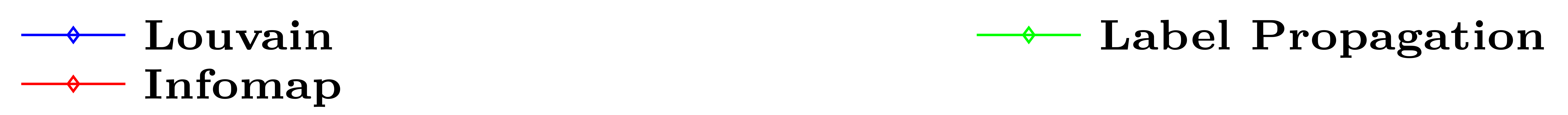}
    \caption{Quantitative comparison between \LP, \LOUVAIN{} and \INFOMAP{} on the \BSDS{} dataset.  
    The results presented use a threshold value of $0.98$ and are shown with radii $\{1,2,5,10\}$. \label{fig:quantitative_metrics}} 
\end{figure}

As one may observe \cref{fig:quantitative_metrics}, considered methods provide relevant results with respect to chosen metrics. 
While all methods result in significant outcomes, \INFOMAP{} outperforms other considered algorithms. To emphasize the impact the radius 
may have on each algorithm, we choose $K=1000$ as the number of computed superpixels. The difference 
observed on metrics is actually similar for all values of superpixels and across datasets. 
An interesting observation is that using a \RPG{10} improves both $\Rec$ and $\EV$ but not $\UE$ while increasing the size of the graph.  
Even if performance is not taken into account in this study, we note that increasing the radius may slow down the computation of communities and 
hence does not seem to be well-suited for this framework. 
As mentioned previously, the set of communities computed by each algorithm shows significant variations that may explain 
the difference in computed metrics. This is also emphasized by a qualitative comparison of each algorithm. 

\begin{figure*}[ht]
    \centering
    \begin{subfigure}[b]{0.025\textwidth}
        \rotatebox{90}{\small\hphantom{aaai}\LOUVAIN}
	\end{subfigure}
    \begin{subfigure}[b]{0.025\textwidth}
        ~
	\end{subfigure}
    \begin{subfigure}[b]{\fullfourone\textwidth}
        \centering\includegraphics[height=2.2cm]{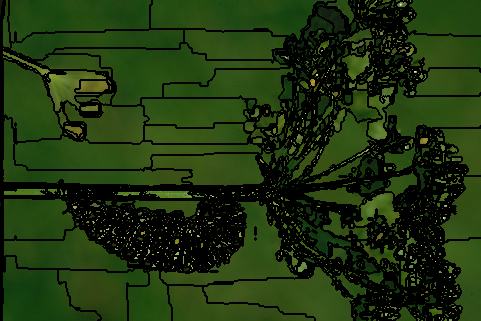}
    \end{subfigure}
	\begin{subfigure}[b]{\fullfourone\textwidth}
        \centering\includegraphics[height=2.2cm]{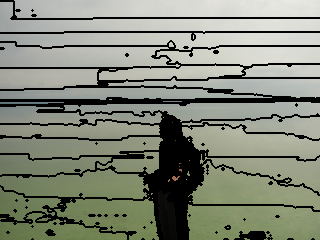}
    \end{subfigure}
	\begin{subfigure}[b]{\fullfourone\textwidth}
        \centering\includegraphics[height=2.2cm]{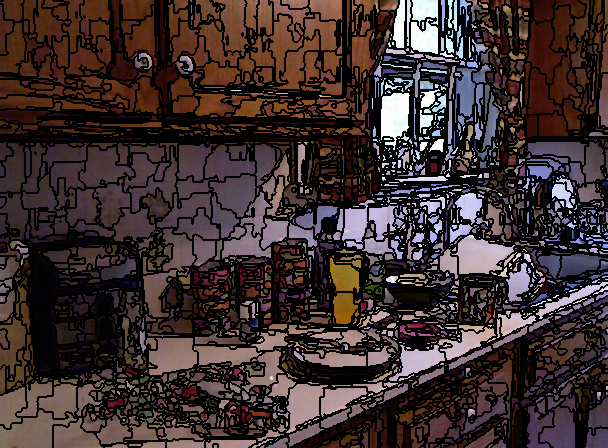}
    \end{subfigure}
	\begin{subfigure}[b]{\fullfourone\textwidth}
        \centering\includegraphics[height=2.2cm]{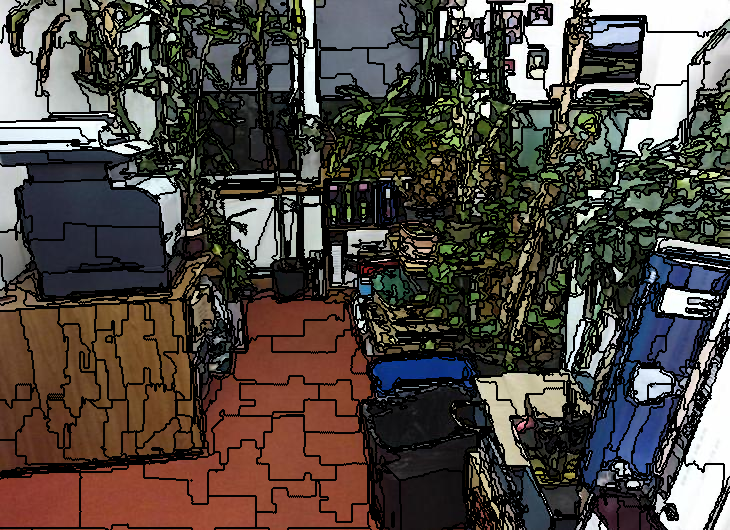}
    \end{subfigure}	\\[4px]
    \begin{subfigure}[b]{0.025\textwidth}
        \rotatebox{90}{\small\hphantom{aaai}\INFOMAP}
    \end{subfigure}
    \begin{subfigure}[b]{0.025\textwidth}
        \rotatebox{90}{\small\hphantom{aaai}$K=2500$}
	\end{subfigure}
    \begin{subfigure}[b]{\fullfourone\textwidth}
        \centering\includegraphics[height=2.2cm]{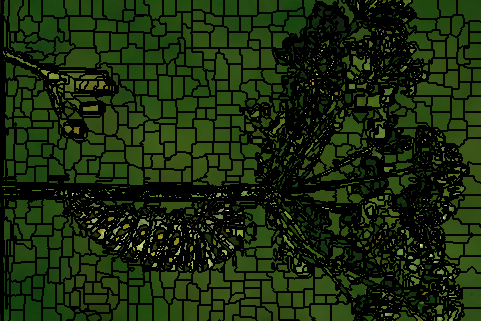}
    \end{subfigure}
	\begin{subfigure}[b]{\fullfourone\textwidth}
        \centering\includegraphics[height=2.2cm]{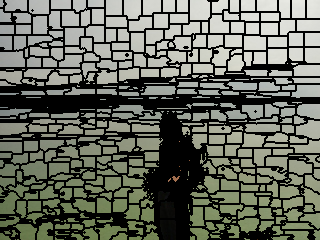}
    \end{subfigure}
	\begin{subfigure}[b]{\fullfourone\textwidth}
        \centering\includegraphics[height=2.2cm]{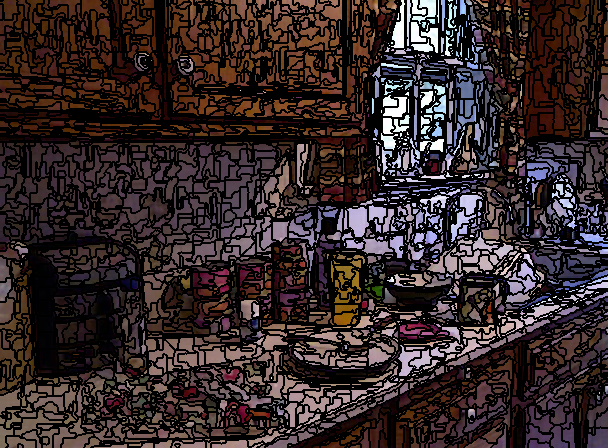}
    \end{subfigure}
	\begin{subfigure}[b]{\fullfourone\textwidth}
        \centering\includegraphics[height=2.2cm]{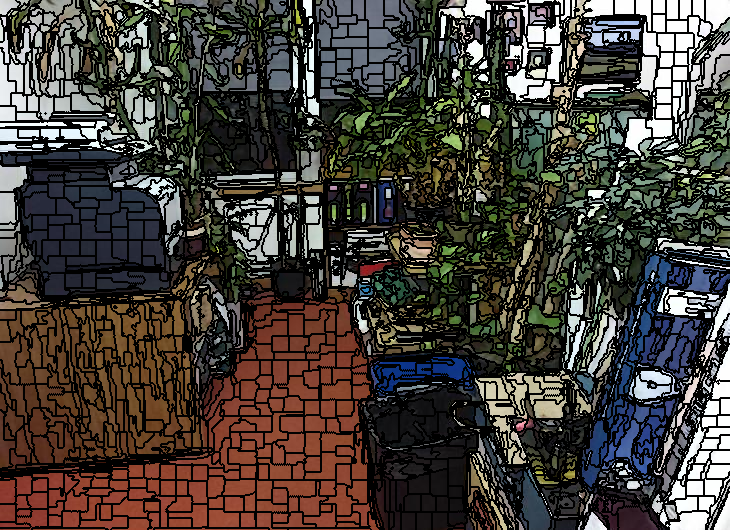}
    \end{subfigure}\\[4px]
    \begin{subfigure}[b]{0.025\textwidth}
        \rotatebox{90}{\small\hphantom{aaai}\texttt{LP}}
	\end{subfigure}
    \begin{subfigure}[b]{0.025\textwidth}
        ~
	\end{subfigure}
    \begin{subfigure}[b]{\fullfourone\textwidth}
        \centering\includegraphics[height=2.2cm]{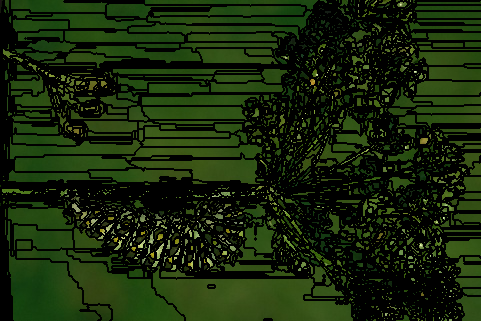}
    \end{subfigure}
	\begin{subfigure}[b]{\fullfourone\textwidth}
        \centering\includegraphics[height=2.2cm]{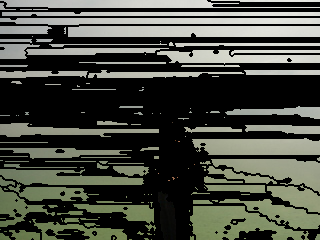}
    \end{subfigure}
	\begin{subfigure}[b]{\fullfourone\textwidth}
        \centering\includegraphics[height=2.2cm]{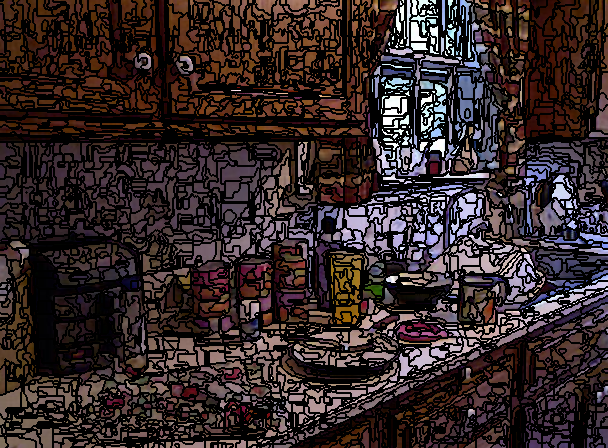}
    \end{subfigure}
	\begin{subfigure}[b]{\fullfourone\textwidth}
        \centering\includegraphics[height=2.2cm]{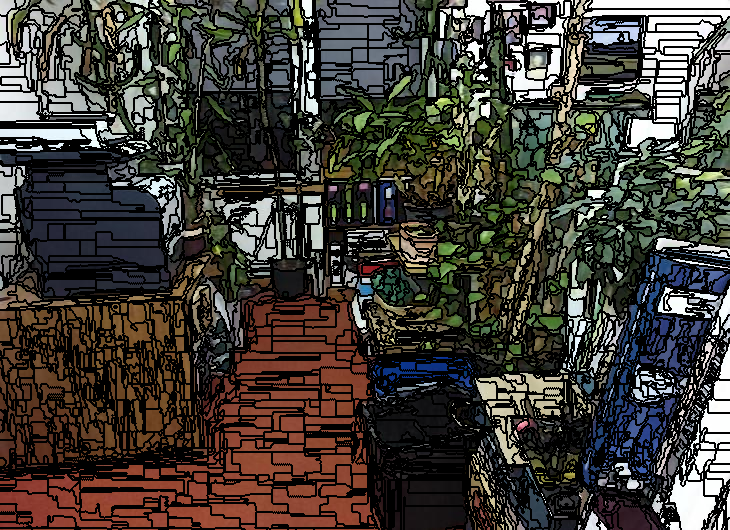}
    \end{subfigure}	\\[4px]
    \medskip
    \centering
    \begin{subfigure}[b]{0.025\textwidth}
        \rotatebox{90}{\small\hphantom{aaai}\LOUVAIN}
	\end{subfigure}
    \begin{subfigure}[b]{0.025\textwidth}
        ~
	\end{subfigure}
	\begin{subfigure}[b]{\fullfourone\textwidth}
        \centering\includegraphics[height=2.2cm]{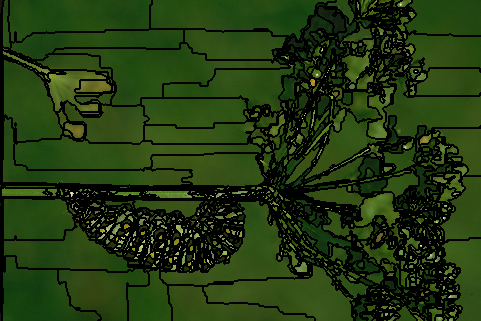}
    \end{subfigure}
	\begin{subfigure}[b]{\fullfourone\textwidth}
        \centering\includegraphics[height=2.2cm]{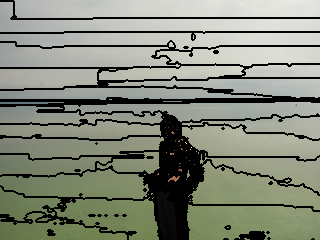}
    \end{subfigure}
	\begin{subfigure}[b]{\fullfourone\textwidth}
        \centering\includegraphics[height=2.2cm]{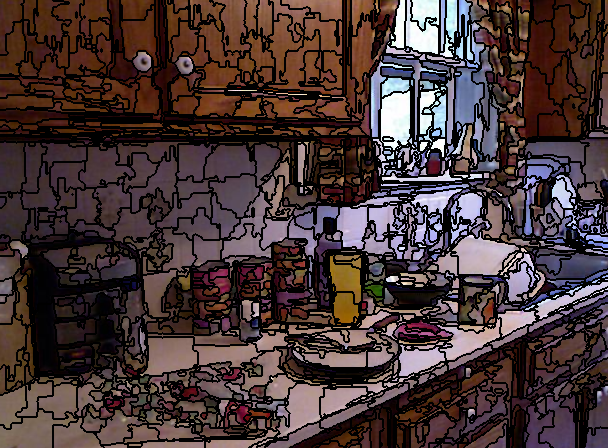}
    \end{subfigure}
	\begin{subfigure}[b]{\fullfourone\textwidth}
        \centering\includegraphics[height=2.2cm]{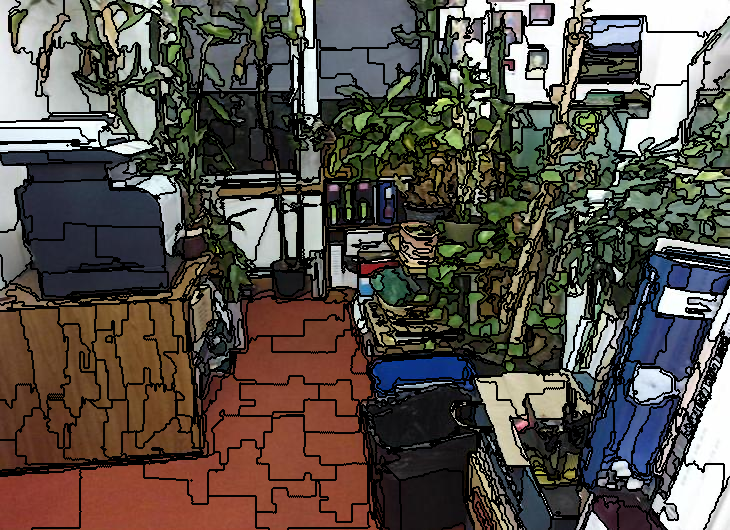}
    \end{subfigure}\\[4px]
    \begin{subfigure}[b]{0.025\textwidth}
        \rotatebox{90}{\small\hphantom{aaai}\INFOMAP}
	\end{subfigure}
    \begin{subfigure}[b]{0.025\textwidth}
        \rotatebox{90}{\small\hphantom{aaai}$K=1000$}
	\end{subfigure}
	\begin{subfigure}[b]{\fullfourone\textwidth}
        \centering\includegraphics[height=2.2cm]{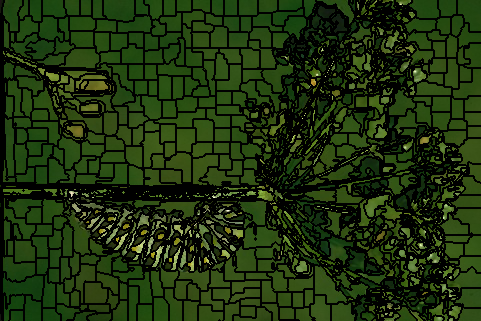}
    \end{subfigure}
	\begin{subfigure}[b]{\fullfourone\textwidth}
        \centering\includegraphics[height=2.2cm]{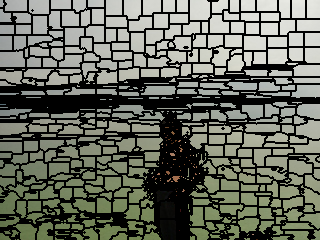}
    \end{subfigure}
	\begin{subfigure}[b]{\fullfourone\textwidth}
        \centering\includegraphics[height=2.2cm]{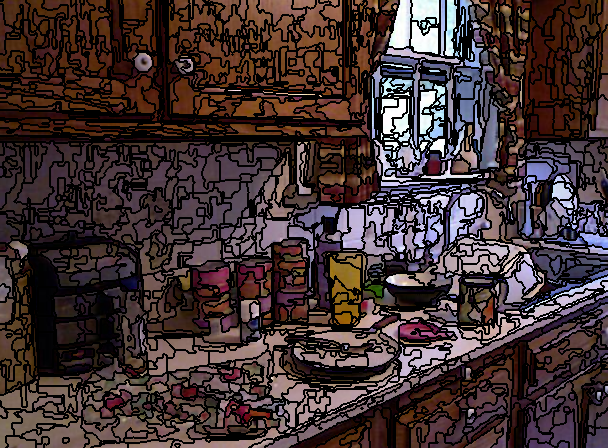}
    \end{subfigure}
	\begin{subfigure}[b]{\fullfourone\textwidth}
        \centering\includegraphics[height=2.2cm]{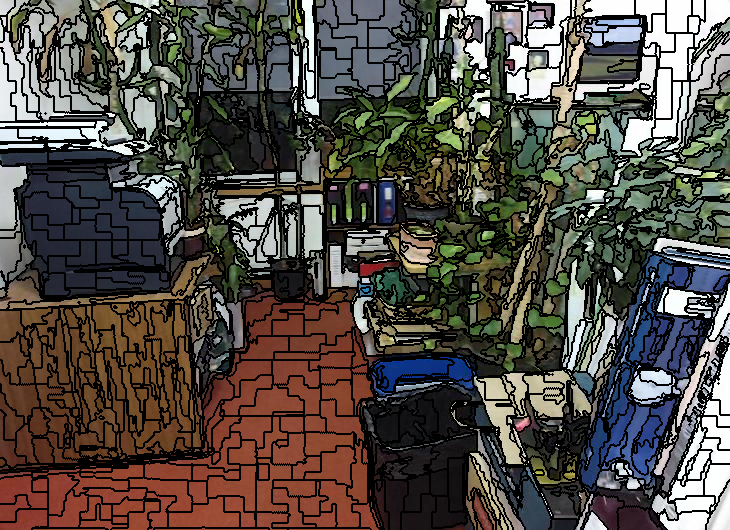}
    \end{subfigure}	\\[4px]
    \begin{subfigure}[b]{0.025\textwidth}
        \rotatebox{90}{\small\hphantom{aaai}\texttt{LP}}
	\end{subfigure}
    \begin{subfigure}[b]{0.025\textwidth}
        ~
	\end{subfigure}
	\begin{subfigure}[b]{\fullfourone\textwidth}
        \centering\includegraphics[height=2.2cm]{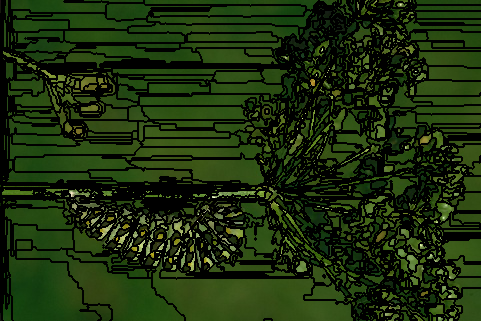}
    \end{subfigure}
	\begin{subfigure}[b]{\fullfourone\textwidth}
        \centering\includegraphics[height=2.2cm]{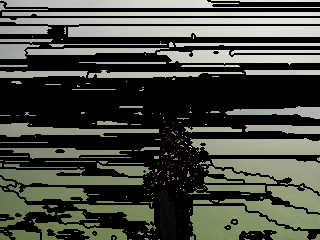}
    \end{subfigure}
	\begin{subfigure}[b]{\fullfourone\textwidth}
        \centering\includegraphics[height=2.2cm]{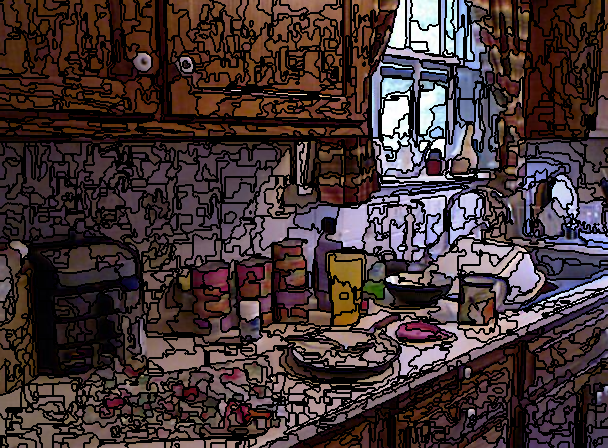}
    \end{subfigure}
	\begin{subfigure}[b]{\fullfourone\textwidth}
        \centering\includegraphics[height=2.2cm]{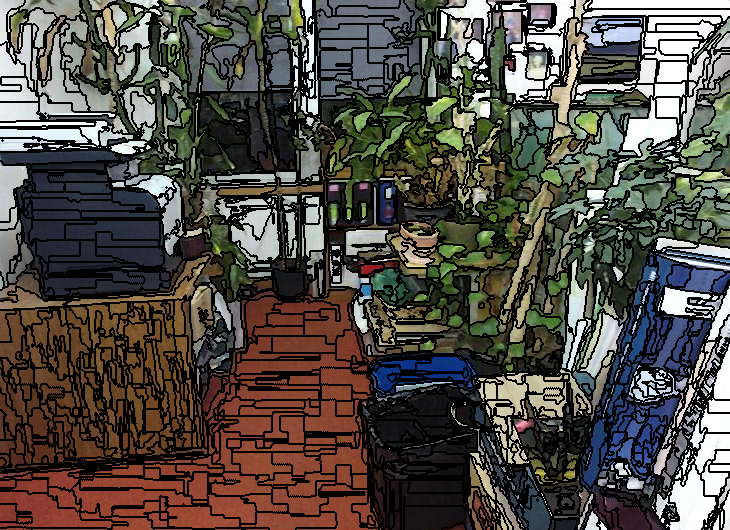}
    \end{subfigure}	\\[4px]
    \medskip
    \centering
    \begin{subfigure}[b]{0.025\textwidth}
        \rotatebox{90}{\small\hphantom{aaai}\LOUVAIN}
	\end{subfigure}
    \begin{subfigure}[b]{0.025\textwidth}
        ~
	\end{subfigure}
    \begin{subfigure}[b]{\fullfourone\textwidth}
        \centering\includegraphics[height=2.2cm]{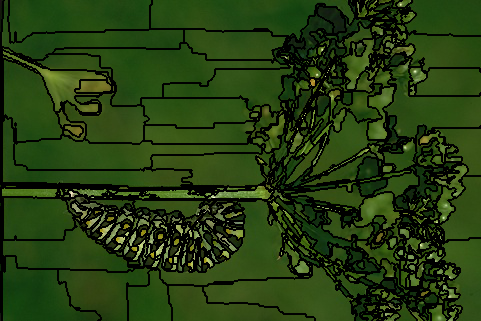}
    \end{subfigure}
	\begin{subfigure}[b]{\fullfourone\textwidth}
        \centering\includegraphics[height=2.2cm]{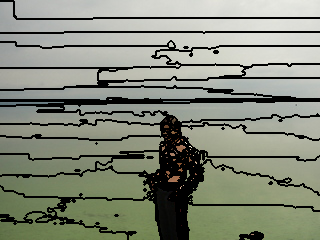}
    \end{subfigure}
	\begin{subfigure}[b]{\fullfourone\textwidth}
        \centering\includegraphics[height=2.2cm]{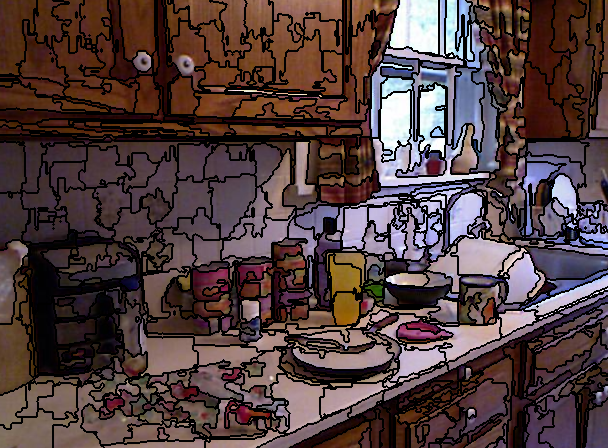}
    \end{subfigure}
	\begin{subfigure}[b]{\fullfourone\textwidth}
        \centering\includegraphics[height=2.2cm]{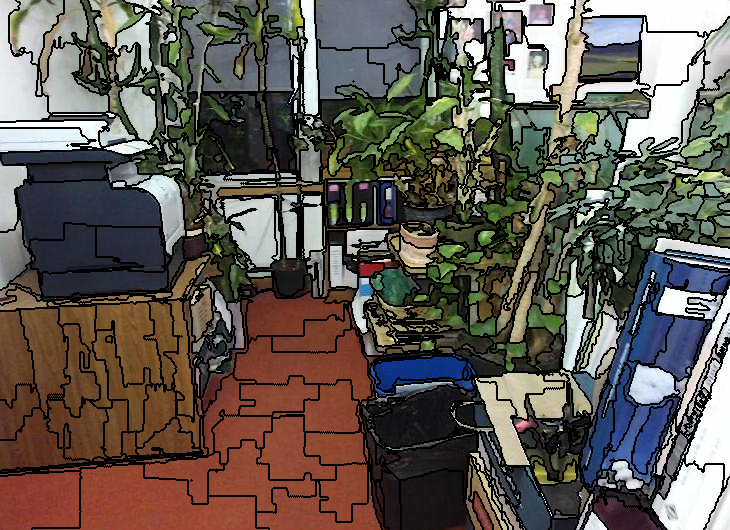}
    \end{subfigure}	\\[4px]
    \begin{subfigure}[b]{0.025\textwidth}
        \rotatebox{90}{\small\hphantom{aaai}\INFOMAP}
	\end{subfigure}
    \begin{subfigure}[b]{0.025\textwidth}
        \rotatebox{90}{\small\hphantom{aaai}$K=400$}
	\end{subfigure}
    \begin{subfigure}[b]{\fullfourone\textwidth}
        \centering\includegraphics[height=2.2cm]{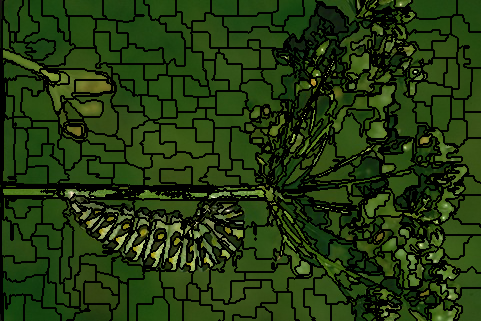}
    \end{subfigure}
	\begin{subfigure}[b]{\fullfourone\textwidth}
        \centering\includegraphics[height=2.2cm]{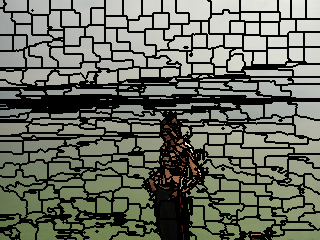}
    \end{subfigure}
	\begin{subfigure}[b]{\fullfourone\textwidth}
        \centering\includegraphics[height=2.2cm]{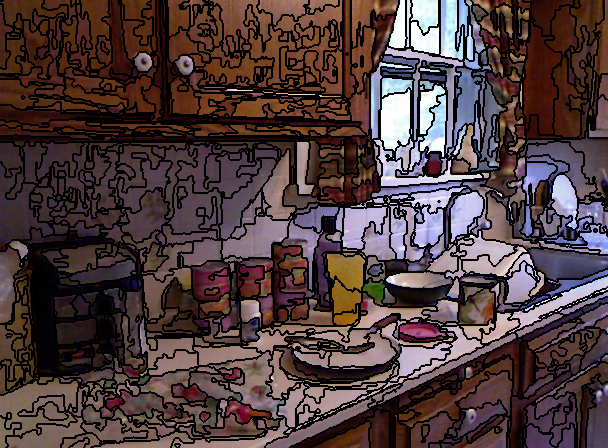}
    \end{subfigure}
	\begin{subfigure}[b]{\fullfourone\textwidth}
        \centering\includegraphics[height=2.2cm]{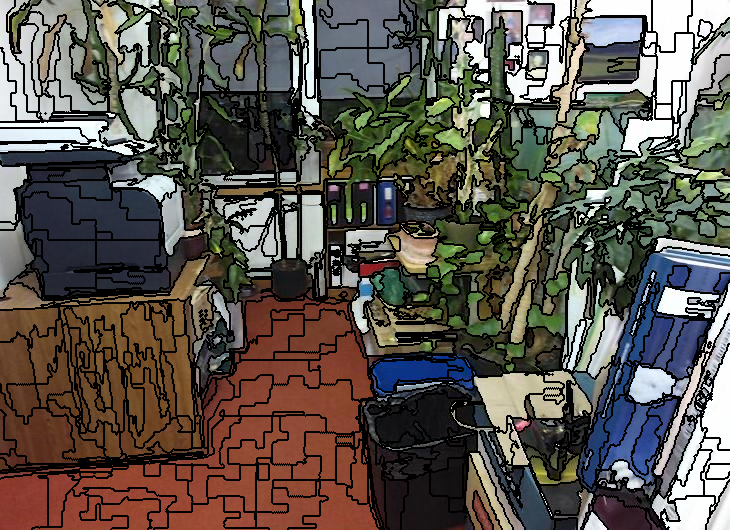}
    \end{subfigure}	\\[4px]
    \begin{subfigure}[b]{0.025\textwidth}
        \rotatebox{90}{\small\hphantom{aaai}\texttt{LP}}
	\end{subfigure}
    \begin{subfigure}[b]{0.025\textwidth}
        ~
	\end{subfigure}
    \begin{subfigure}[b]{\fullfourone\textwidth}
        \centering\includegraphics[height=2.2cm]{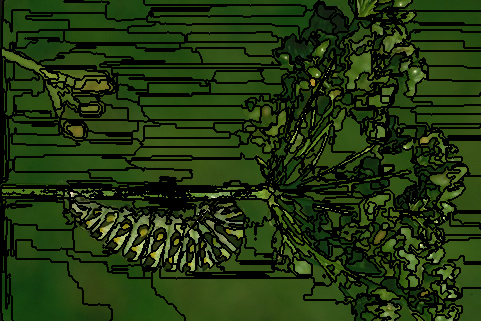}
    \end{subfigure}
	\begin{subfigure}[b]{\fullfourone\textwidth}
        \centering\includegraphics[height=2.2cm]{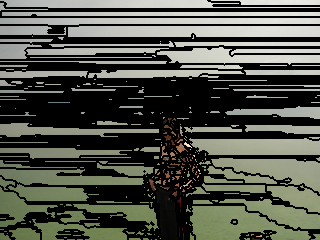}
    \end{subfigure}
	\begin{subfigure}[b]{\fullfourone\textwidth}
        \centering\includegraphics[height=2.2cm]{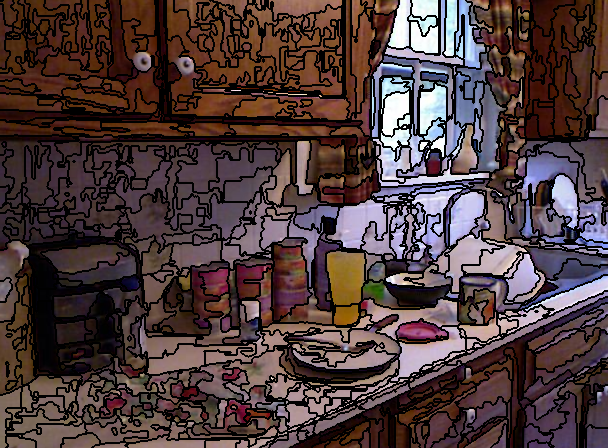}
    \end{subfigure}
	\begin{subfigure}[b]{\fullfourone\textwidth}
        \centering\includegraphics[height=2.2cm]{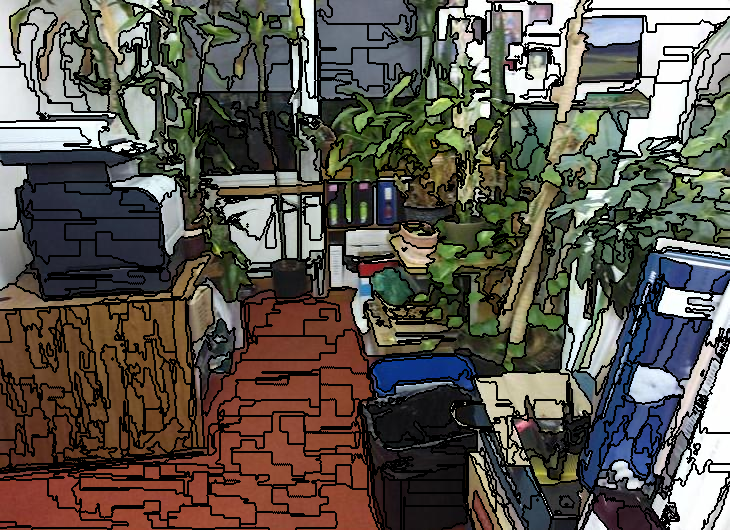}
    \end{subfigure}	
    \caption{Qualitative comparison of outputted superpixels for selected datasets and 
    community detection algorithms. \label{fig:qualitative_communities}}
\end{figure*}

\paragraph{Qualitative difference between algorithms}
\cref{fig:qualitative_communities} shows that \INFOMAP{} tends to exhibit more compact and regular superpixels. 
In particular, on images with a clearly identified background both \LP{} and \LOUVAIN{} tend to group the background into 
a few large superpixels. 
This is mainly due to the ability of \INFOMAP{} to produce smaller communities (recall \cref{table:communities_stats}), 
which hence results in a regular initial oversegmentation for which the merging procedure yields relevant results.

\section{Conclusion}
\label{sec:conclusion}

In this work we analyzed the relevance of superpixels computed through community detection algorithms and a merging procedure 
on a simple graph extracted from the image at hand. As mentioned \cref{sec:introduction} this approach has previously been used 
to compute \emph{undersegmentation} in several works~\cite{CD_AMB14,CD_LBG+17,CD_HvGB+12,CD_BL16,CD_BAV11,CD_MEC19,CD_NGC20,CD_LW14} but, to the best of our knowledge, its use for the computation of superpixels has 
been neglected so far, the work of Liu \etal~\cite{LDG+22} being the only notable exception.  
We hence enlight the relevance of this approach by providing both qualitative and quantitative results w.r.t. state-of-the-art methods. One 
interesting property of this approach is its ability to evolve in time: novel community detection algorithms are frequently 
introduced (see \eg~\cite{FN22}) and any improvement may also improve this approach. Notice that in this work we chose to rely 
on three well-studied and widely used algorithms to ensure the relevance of presented results. 
Hence as one of the most straightforward extension one may consider the analysis of more recent 
community detection algorithms in order to better understand the impact of such a choice. In particular it would be interesting 
to use a community detection algorithm that allows control on the number of computed communities. 
Another approach is to consider the graph extracted from the image. We here made the choice to work on a very simple graph, 
in particular using a weight function that only relies on color similarities. It would be interesting to 
study the impact of a more intricate similarity measure, for instance accounting for the histogram of oriented gradients as 
in~\cite{CD_NGC20} or with a trade-off between feature similarity and Euclidean distance as in~\cite{ZGZ21}. 
We would like to mention that we also conducted experiments without weighting the graph, resulting in less relevant results. 
This seems to indicate that choosing a different weight function may have some impact on the results. 
Finally, the choices made for the merging procedure may result in different outcomes. We studied a basic strategy that 
provide relevant results, and it could be interesting to study further the impact of such a procedure. We note here that we tried 
another approach that merged really small regions first and then neighboring remaining regions if their similarity is above 
the similarity threshold $\rho=0.98$. Quite expectedly, this approach provided similar results and does not seem to lead to any improvement. 

\paragraph{Acknowledgments} The author is deeply grateful to David Stutz for fruitful discussion and for pointing out the 
availability of the \LaTeX{} \href{https://github.com/davidstutz/cviu2018-superpixels}{source} for the paper Superpixels: An evaluation of the state-of-the-art~\cite{SHL18}. 

\clearpage
\bibliography{bibliography.bib}

\end{document}